\documentclass{article}

\usepackage[a4paper, bindingoffset=0.2in,
            left=0.9in,right=0.9in, top=1in, bottom=1in,
            footskip=0.25in]{geometry}

\usepackage[utf8]{inputenc} % allow utf-8 input
\usepackage[T1]{fontenc}    % use 8-bit T1 fonts
\usepackage{hyperref}       % hyperlinks
\usepackage{url}            % simple URL typesetting
\usepackage{booktabs}       % professional-quality tables
\usepackage{amsfonts}       % blackboard math symbols
\usepackage{nicefrac}       % compact symbols for 1/2, etc.
\usepackage{microtype}      % microtypography
\usepackage{xcolor}         % colors
\usepackage{algorithm}
\usepackage{algorithmic}
\usepackage{gensymb}
\usepackage{subcaption}
\usepackage{multicol}
\usepackage{multirow}
\usepackage{graphicx}
\usepackage{amsmath}
\usepackage{amsthm}

\def\1{\bm{1}}

\usepackage{bm}

\def\rvtheta{{\bm{\theta}}}
\def\rveta{{\bm{\eta}}}

\def\rvx{{\mathbf{x}}}

\DeclareMathAlphabet{\mathsfit}{\encodingdefault}{\sfdefault}{m}{sl}
\SetMathAlphabet{\mathsfit}{bold}{\encodingdefault}{\sfdefault}{bx}{n}

\def\gB{{\mathcal{B}}}

\def\gD{{\mathcal{D}}}

\newcommand{\argmin}{arg\,min}
\renewcommand{\algorithmicrequire}{ \textbf{Input:}} %Use Input in the format of Algorithm
\renewcommand{\algorithmicensure}{ \textbf{Output:}}
\usepackage[capitalize,noabbrev]{cleveref}

\def\name{{ATAS}}

\DeclareRobustCommand\onedot{\futurelet\@let@token\@onedot}
\def\@onedot{\ifx\@let@token.\else.\null\fi\xspace}

\makeatother
\newcommand{\change}[1]{{\color{black}{#1}}}

\title{Fast Adversarial Training with Adaptive Step Size}

\author{
  Zhichao Huang$^\text{1}$, Yanbo Fan$^\text{2}$, Chen Liu$^\text{3}$, Weizhong Zhang$^\text{1}$,
  Yong Zhang$^\text{2}$ \\ Mathieu Salzmann$^\text{3}$, Sabine S\"usstrunk$^\text{3}$, Jue Wang$^\text{2}$  \\
  $^1$The Hong Kong University of Science and Technology\qquad$^2$Tencent AI Lab\\
  $^3$École Polytechnique Fédérale de Lausanne
}

\date{}
\begin{document}

\maketitle

\begin{abstract}

    While adversarial training and its variants have shown to be the most effective algorithms to defend against adversarial attacks, their extremely slow training process makes it hard to scale to large datasets like ImageNet. The key idea of recent works to accelerate adversarial training is to substitute multi-step attacks (e.g., PGD) with single-step attacks (e.g., FGSM). However, these single-step methods suffer from catastrophic overfitting, where the accuracy against PGD attack suddenly drops to nearly 0\% during training, destroying the robustness of the networks. In this work, we study the phenomenon from the perspective of training instances. We show that catastrophic overfitting is instance-dependent and fitting instances with larger gradient norm is more likely to cause catastrophic overfitting. Based on our findings, we propose a simple but effective method, \textit{Adversarial Training with Adaptive Step size (ATAS)}. ATAS learns an instance-wise adaptive step size that is inversely proportional to its gradient norm. The theoretical analysis shows that {\name} converges faster than the commonly adopted non-adaptive counterparts. Empirically, {\name} consistently mitigates catastrophic overfitting and achieves higher robust accuracy on CIFAR10, CIFAR100 and ImageNet when evaluated on various adversarial budgets.
\end{abstract}

\section{Introduction}

Adversarial examples \cite{szegedy2013intriguing} cause serious safety concerns in deploying deep learning models. In order to defend against adversarial attacks, many approaches have been proposed \cite{guo2017countering, liao2018defense, at, trades}. 
Among them, adversarial training and its variants \cite{at, mart, trades} have been recognized as the most effective defense mechanism. 
Adversarial training (AT) is generally formulated as a minimax  problem
\begin{equation}
    \min_\rvtheta \max_{\rvx_i^* \in \gB_p(\rvx_i, \varepsilon)} \frac{1}{n} \sum_{i=1}^n \ell(\rvx_i^*, y_i; \rvtheta)\;
    \label{eqn:at},
\end{equation}
where $\gD=(\rvx_i,y_i)_{i=1}^n$ is the training set and $\ell(\rvx,y; \rvtheta)$ is the loss function parametrized by $\rvtheta$. $\gB_p(\rvx_i, \varepsilon)$ represents a $L_p$ norm ball centered at $\rvx_i$ with radius $\varepsilon$. 
AT in \Cref{eqn:at} boosts the adversarial robustness by adopting adversarial examples generated in the inner maximization.

Despite the effectiveness of AT, solving the inner maximization requires multiple steps of projected gradient descent (PGD) \cite{at, overfitting}. Therefore, AT is much slower than vanilla training (\textit{e.g.}, 10 times longer training time for AT in \cite{overfitting}),
making it challenging to scale AT to large datasets such as ImageNet. 

Currently, the typical solution to accelerate AT is to substitute multi-step attacks (e.g., PGD) with single-step attacks (e.g., FGSM). Several works have been proposed following this direction, including FGSM-RS \cite{fastat}, ATTA \cite{atta} \textit{etc}. These methods achieve the best robust accuracy for fast AT. However, recent works \cite{fgsmga, kim2020understanding} demonstrate that the single-step method suffers from catastrophic overfitting, where the model's robustness against PGD attack suddenly drops to nearly 0\% while the robust accuracy against FGSM attack rapidly increases \cite{fastat}. This will completely destroy the robustness of the networks. It is worth noting that catastrophic overfitting is different from robust overfitting mentioned in \cite{overfitting}. The latter one refers to the generalization gap between training and test data while catastrophic overfitting means the overfitting to a specific type of attack that is irrelevant to the training and test set. 
Some works have been proposed to understand and alleviate the catastrophic overfitting \cite{fgsmga, kim2020understanding}.
However, their solutions significantly increase the training time.
For example, the gradient align regularizer $\mathbb{E}_{\rveta \sim \mathcal{U}\left([-\varepsilon, \varepsilon]^{d}\right)}\left[1-\cos \left(\nabla_{\rvx} \ell(\rvx, y ; \rvtheta), \nabla_{\rvx} \ell(\rvx+\rveta, y ; \rvtheta)\right)\right]$ in \cite{fgsmga} requires calculating the second order gradient and it is still 5 times slower than vanilla training. And \cite{kim2020understanding} needs to check several points within the $\ell_p$ norm ball, which needs several forward propagation and is still about 4 times slower than vanilla training. Therefore, existing methods are still unsatisfactory in terms of both training efficiency and robust performance.

In this work, we analyze catastrophic overfitting from the perspective of training instances. By taking the gradient norm as an indicator, we find that different training instances have different probabilities of causing catastrophic overfitting. Instances with large gradient norm are more sensitive to the adversarial noise and their loss landscape is less smooth. Thus, fitting them with FGSM is more likely to distort the loss landscape, resulting in catastrophic overfitting.

Furthermore, catastrophic overfitting is closely related to the optimization process of the inner maximization, \textit{e.g.}, the setting of step size. When catastrophic overfitting does not occur, the larger step size leads to a stronger attack and thus strengthens the robustness of the network \cite{fastat}. On the other side, a larger step size is more likely to cause catastrophic overfitting in the training process \cite{fgsmga, fastat}. 
Based on these findings, we propose \textit{Adversarial Training with Adaptive Step size ({\name})}, an simple but effective fast AT method that uses the previous initialization in ATTA \cite{atta} and takes the step size of the inner maximization inversely proportional to the input gradient norm. Instances with large gradient norm are given a small step size to prevent catastrophic overfitting. By contrast, instances with small gradient norms will have large step sizes to improve the strength of the attack. 

We theoretically analyze the convergence of {\name} and prove that it converges faster than the non-adaptive counterpart, which is commonly adopted in existing works \cite{atta}, especially when the distribution of the input gradient norm is long-tailed. Empirically, We evaluate {\name} on CIFAR10, CIFAR100 \cite{cifar10} and ImageNet \cite{deng2009imagenet} with different network architectures and adversarial budgets, showing that {\name} mitigates catastrophic overfitting and achieves higher robust accuracy under various attacks including PGD10, PGD50 \cite{at} and AutoAttack \cite{autoattack}. 

Our contributions are summarized as follows:
1) To the best of our knowledge, we are the first to analyze catastrophic overfitting from the perspective of training instances, and demonstrate that instances with large input gradient norms are more likely to cause catastrophic overfitting.
2) Based on our findings, we propose a new algorithm, {\name}, which takes the step size of the inner maximization to be inversely proportional to the input gradient norm in order to prevents catastrophic overfitting and maintain the strength of the attack. 
3) Theoretically, we prove that {\name} converges faster than its non-adaptive counterpart.
4) Empirically, we conduct extensive experiments to evaluate {\name} on different datasets, network architectures and adversarial budgets, showing that {\name} consistently improves the robust accuracy and mitigates catastrophic overfitting.

\section{Background and Related Work}
 
\subsection{Adversarial Examples.}

Adversarial examples are first discussed in \cite{szegedy2013intriguing}, where a small  perturbation of the input significantly changes the prediction. Adversarial examples can be generated using the gradient of the input $\rvx$. Fast Gradient Signed Method (FGSM) \cite{fgsm} approximates the loss function $\ell(\rvx, y; \rvtheta)$ with the first order Taylor expansion so that adversarial examples can be generated with one step of projected gradient
$
    \rvx^{\text{FGSM}} = \rvx + \varepsilon \cdot \text{sgn} (\nabla_{\rvx}\ell(\rvx, y; \rvtheta)))\;,
    % \label{eqn:fgsm}
$
where $\varepsilon$ is the adversarial budget. Projected Gradient Descent (PGD) \cite{at} extends FGSM to multiple steps to strengthen the attack. With a step size $\alpha$, the adversarial example at the $t$-th step is
$
    \rvx^{t+1} = \Pi_{\gB_p(\rvx, \varepsilon)}[\rvx^{t} + \alpha \cdot \text{sgn} (\nabla_{\rvx^{t}}\ell(\rvx^{t}, y; \rvtheta))]\;,
    % \label{eqn:pgd}
$
where $\Pi_{\gB_p(\rvx, \varepsilon)}$ means the projection onto $\gB_p(\rvx, \varepsilon)$. Several stronger attacks are proposed to reliably evaluate the models' robustness \cite{squareattack, fab, autoattack}. Among them, Autoattack \cite{autoattack} stands out as the strongest attack.

While many algorithms \cite{guo2017countering, liao2018defense, at, song2017pixeldefend, mart, trades} have been proposed to defend against adversarial attacks, adversarial training and its variants \cite{at, mart,trades} are shown to be the most effective methods to train a truly robust network. Adversarial training can be formulated as a minimax problem in  \Cref{eqn:at}. Finding solutions of the minimax optimization has been a major endeavor in mathematics and computer science \cite{bacsar1998dynamic, roughgarden2010algorithmic}. Theoretically, the well-known Stochastic Gradient Descent Ascent (SGDA) algorithm finds an $\varepsilon$-approximate stationary point in $\mathcal{O}(1/\varepsilon^2)$ iterations with averaging for convex-concave games \cite{mokhtari2020unified}. However, it is not appropriate to formulate the optimization of AT as SGDA or SGDmax \cite{sgda}, since it only updates a part of the coordinates in $\rvx=[x_1, x_2, \cdots x_n]$ for the maximization. The inner maximization actually corresponds to the stochastic block coordinate ascent.
Empirically, the neural network is non-concave with respect to the input, so perfectly solving the inner maximization is NP-hard. It is usually approximated by a strong attack like PGD \cite{at}, which needs multiple steps of the calculation the gradients. Therefore, adversarial training is much slower than vanilla training.

\subsection{Fast Adversarial Training.}
FreeAT \cite{freeat} first proposes a fast AT method by simultaneously optimizing the model's parameter and the adversarial perturbations by batch replaying.
YOPO \cite{yopo} adopts a similar strategy to optimize the adversarial loss function. Later on, single-step methods are shown to be more effective than FreeAT and YOPO \cite{fastat}. FGSM with Random Start (FGSM-RS) can be used to generate adversarial perturbations in one step to train a robust network if the hyperparameters are carefully tuned \cite{fastat}. ATTA \cite{atta} utilizes the transferability of adversarial examples between epochs, using adversarial example of the previous epoch as the initialization, optimizing the model parameters with
\begin{equation}
\begin{aligned}
    \rvx_i^{j} &= \Pi_{\gB_p(\rvx_i, \varepsilon)}[\rvx_i^{j-1} + \alpha \cdot \text{sgn} (\nabla_{\rvx_i^{j-1}}\ell(\rvx_i^{j-1}, y; \rvtheta))] \\
    \rvtheta & = \rvtheta - \eta \nabla_{\rvtheta}\ell(\rvx_i^j, y; \rvtheta))\;,
\end{aligned}
\end{equation}
where $\rvx_i^j$ means the adversarial examples generated for the $i$-th instance $\rvx_i$ at the $j$-th epoch. ATTA shows comparable robust accuracy with FGSM-RS. SLAT \cite{slat} perturbs both inputs and the latents simultaneously with FGSM, ensuring more reliable performance.

As mentioned above, these single-step methods suffer from \textit{catastrophic overfitting}, meaning the robustness against PGD attack suddenly drops to nearly 0\% while the robust accuracy against FGSM attack rapidly increases.
In order to prevent catastrophic overfitting, FGSM-GA \cite{fgsmga} adds a regularizer that aligns the direction of the input gradient. Another work \cite{kim2020understanding} studies the phenomenon from of the perspective of loss landscape, finding that catastrophic overfitting is a result of highly distorted loss surface. It proposes a new algorithm to resolve catastrophic overfitting by checking the loss value along the direction of the gradient. However, both algorithms require much more computation than FGSM-RS \cite{fastat} and ATTA \cite{atta}. Compared with these works, we study catastrophic overfitting from the perspective of training instances and show that using adaptive step sizes in single-step methods prevents catastrophic overfitting. Our method achieves better performance with negligible computational overhead.
Adaptive step sizes have been widely used in training neural networks such as AdaGrad \cite{adagrad}, RMSProp \cite{rmsprop} and ADAM \cite{fang2019convergence, adam, amsgrad}. However, our motivation is different, and to the best of our knowledge, we are the first to introduce the adaptive step size in fast AT. 

\section{Motivation}
\label{sec:motivation}

Catastrophic overfitting is interpreted as a result of highly distorted loss landscapes of the input \cite{kim2020understanding}. 
For examples, FGSM-RS \cite{fastat} uses large step sizes in the inner maximization to generate adversarial examples. It may only minimize the classification loss near the boundary of the adversarial budget, while the loss inside the adversarial budget may increase, leading to a highly distorted loss landscapes.
 
Recalling that different inputs have different loss landscape, they may result in different probabilities of causing catastrophic overfitting.  Instances with large gradient norms are more sensitive to the adversarial noise. Thus, the network may simply minimize the loss on the FGSM-perturbed examples near the boundary instead of the whole space within the adversarial budget. This leads to highly distorted loss landscapes and catastrophic overfitting. The following experiments verify our hypothesis of catastrophic overfitting in FGSM-RS. The results of ATTA \cite{atta} are deferred to the \Cref{sec:addexp_atta}.

\begin{figure}[t]
	\begin{subfigure}[t]{0.49\linewidth}
		\centering
		  %  \vspace{-0.2cm}
            \includegraphics[width=\linewidth]{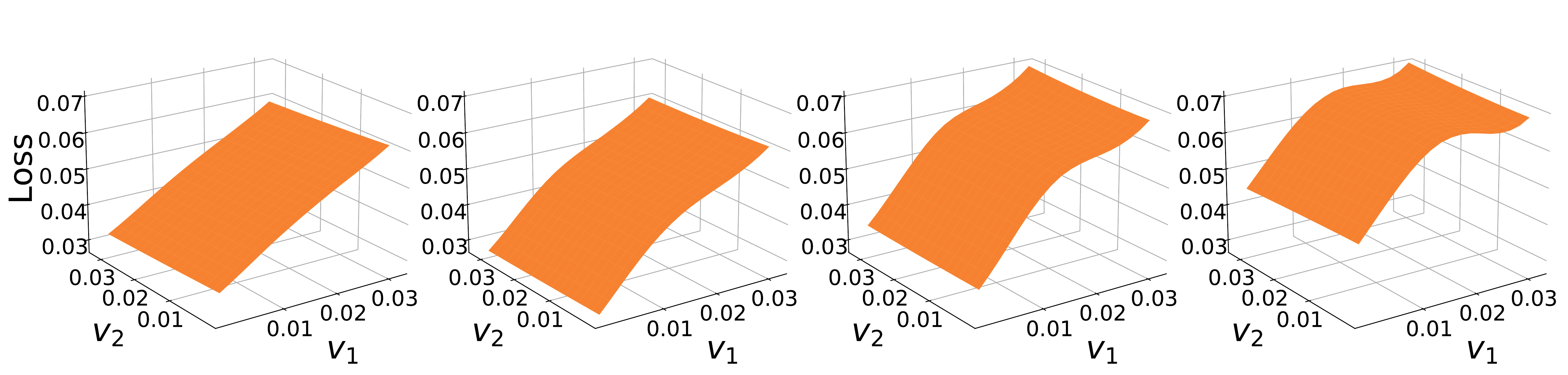}
            \caption{$\gD_1^1$ (Instances with smallest 10\% gradient norm)}
            \label{fig:loss_gdnorm_easy}
    \end{subfigure}
	\begin{subfigure}[t]{0.49\linewidth}
		\centering
            \includegraphics[width=\linewidth]{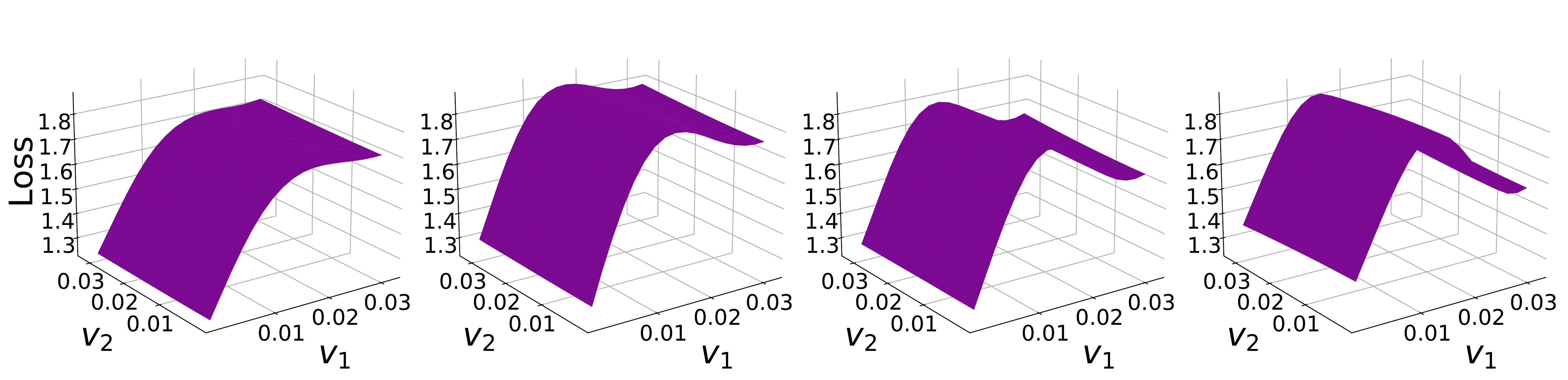}
            \caption{$\gD_{10}^{10}$ (Instances with largest 10\% gradient norm)}
            \label{fig:loss_gdnorm_hard}
    \end{subfigure}
    \vspace{-0.2cm}
    \caption{The loss surface of the subsets $\gD_1^1$ and $\gD_{10}^{10}$. We average the loss of the instances from each subset. $v_1$ is the direction of adversarial noise and $v_2$ is a random direction. Figures from left to right plot the loss surface as the training step increases and each column of (a) and (b) corresponds to the same step of FGSM-RS.}
    \label{fig:loss_co}
    \vspace{-0.5cm}
\end{figure}

\begin{figure*}[t]
    \centering
    \includegraphics[width=\linewidth]{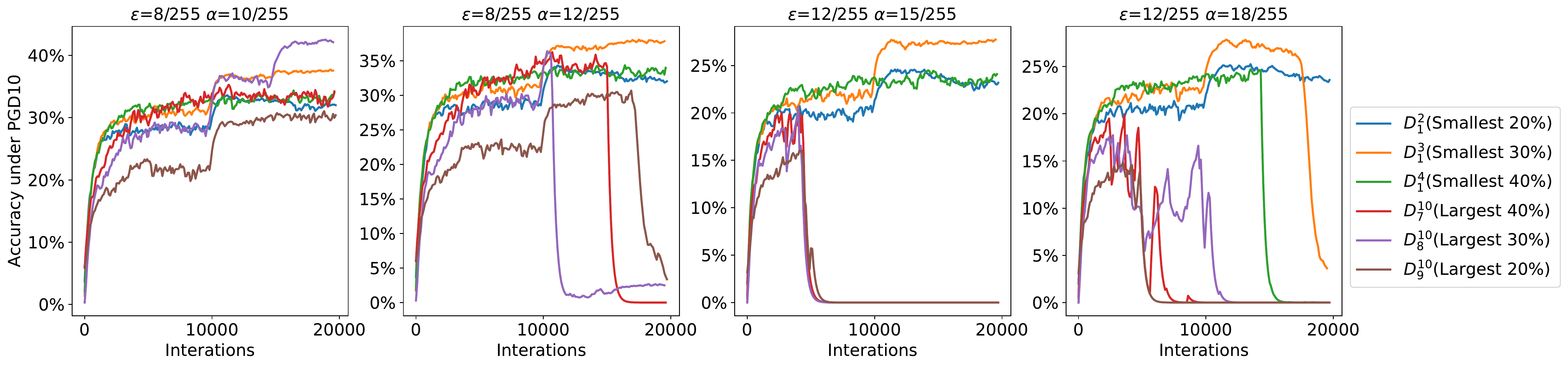}
    \vspace{-0.5cm}
    \caption{The robust training accuracy curve of FGSM-RS trained on different subsets of CIFAR10. The adversarial budgets and the step sizes are shown on top of each figure. The sudden decrease in accuracy indicates catastrophic overfitting. }
    \label{fig:co}
    \vspace{-0.6cm}
\end{figure*}

\noindent\textbf{Metrics of Input Gradient Norm.} To verify the hypothesis that instances with large gradient norms cause catastrophic overfitting, we divide the training instances into different subsets according to their gradient norms. Following the grouping method in \cite{easyhard}, we also average the gradient norm across the training process to reduce the randomness. Formally speaking, we perform FGSM-RS to train a ResNet-18 (RN-18) on CIFAR10 for $N=30$ epochs with $\varepsilon=8/255$ and step size $\alpha = 10/255$. \change{And catastrophic overfitting does not happen in this case.} The average gradient norm 
$
    GN(\rvx_i) = \frac{1}{N} \sum_{j=1}^{N} \|\nabla_{\tilde{\rvx}_i^j} \ell(\tilde{\rvx}_i^j, y_i; \rvtheta)\|_2\
$,
where $\tilde{\rvx}_i^j$ is the random initialization of $\rvx_i$ at the $j$-th epoch. 
We sort $\rvx_i$ according to $GN(\rvx_i)$ and define  
$\text{rank}(\rvx_i) = \frac{1}{n}\sum_{j=1}^n 1(GN(\rvx_j) < GN(\rvx_i))$
as the fraction of instances with smaller average gradient norm than $\rvx_i$. We divide the subsets according to $\text{rank}(\rvx_i)$:
$\gD_i^j = \{\rvx_k| \frac{10(i-1)}{n} \le \text{rank}(\rvx_k) < \frac{10j}{n}\}.$ The classes of each subset is balanced. The maximum and minimum proportion of one class in all subsets is 10.86\% and 8.98\% in CIFAR10.

\noindent\textbf{Loss Landscape.} We train a new RN-18 using FGSM-RS and enlarge the step size to $\alpha=14/255$ to cause catastrophic overfitting.  \Cref{fig:loss_co} shows the loss surface of the subsets with the smallest ($\gD_1^1$) and the largest gradient norm ($\gD_{10}^{10})$ when the catastrophic overfitting happens. $\gD_{10}^{10}$ first exhibits the catastrophic overfitting, where the loss surface of the input gets highly distorted and the loss function reaches its highest value in the middle of the adversarial budget. By contrast, the loss surface of $\gD_1^1$ is less distorted.  \Cref{fig:loss_co} infers that the subsets with large gradient norm are more likely to suffer from catastrophic overfitting.

\noindent\textbf{Training with Different Subsets.} We perform FGSM-RS on different subsets of CIFAR10 with different adversarial budgets $\varepsilon$ and step size $\alpha$ to show that fitting examples with larger gradient norm is more likely to cause catastrophic overfitting. We train the RN-18 on instances with small gradient norm $\gD_1^2$, $\gD_1^3$, $\gD_1^4$ and instances with large gradient norm $\gD_7^{10}$, $\gD_8^{10}$, $\gD_9^{10}$. While different subsets contain different number of instances, we keep the number of the training iterations the same for fair comparison.  
In \Cref{fig:co}, we show the robust accuracy of the whole training set under PGD-10. For $\varepsilon=8/255$ with $\alpha=10/255$, the models trained with all subsets do not exhibit catastrophic overfitting. However, as the step size $\alpha$ increases, subsets with large norms first exhibit catastrophic overfitting, while catastrophic overfitting is less likely to occur in the model trained with the subsets of small gradient norm.  \change{The figure shows 1) for each subset, catastrophic overfitting is more likely to occur when increasing the step size; 2) for a fixed step size, catastrophic overfitting is less likely to happen for subset with small gradient norm.
}

\begin{figure}[t]
    \centering
	\begin{subfigure}[t]{0.25\linewidth}
	    \centering
        \includegraphics[width=\linewidth]{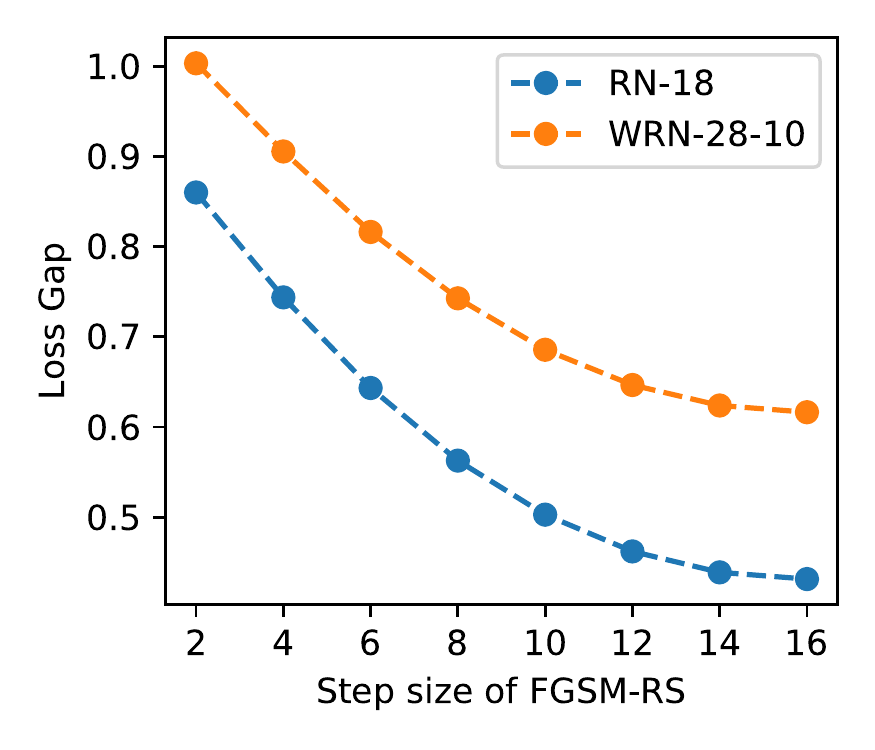}
        \captionsetup{font=normalsize}
                \vspace{-0.6cm}

        \caption{}
        \label{fig:loss_gap}
    \end{subfigure}
	\begin{subfigure}[t]{0.25\linewidth}
	    \centering
        \includegraphics[width=\linewidth]{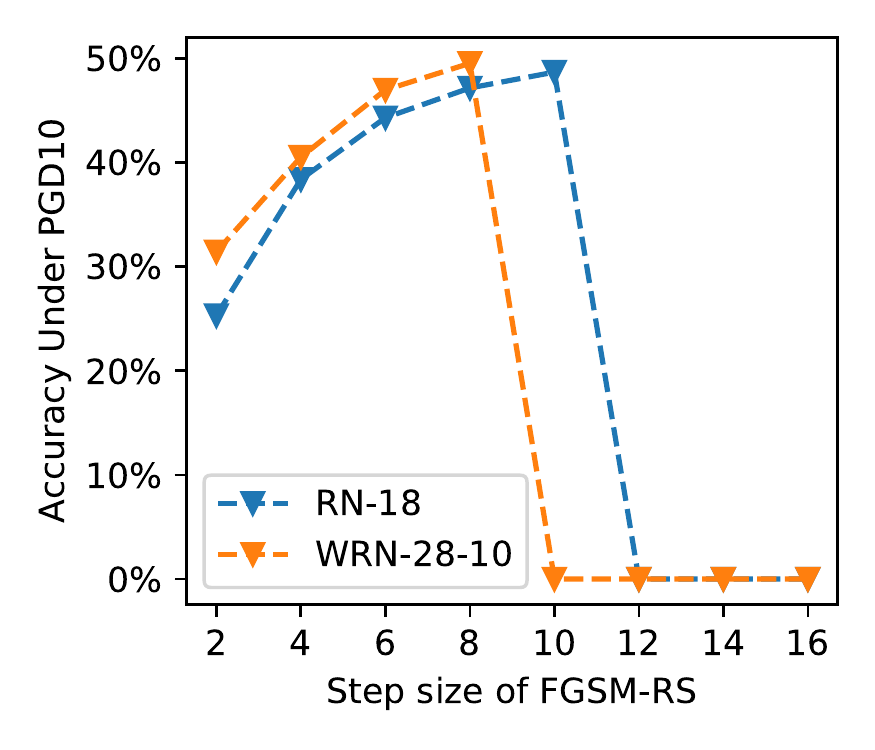}
        \captionsetup{font=normalsize}
                \vspace{-0.6cm}

        \caption{}
        \label{fig:accuracy_stepsize}
    \end{subfigure}
	\begin{subfigure}[t]{0.45\linewidth}
        \centering
        \includegraphics[width=\linewidth]{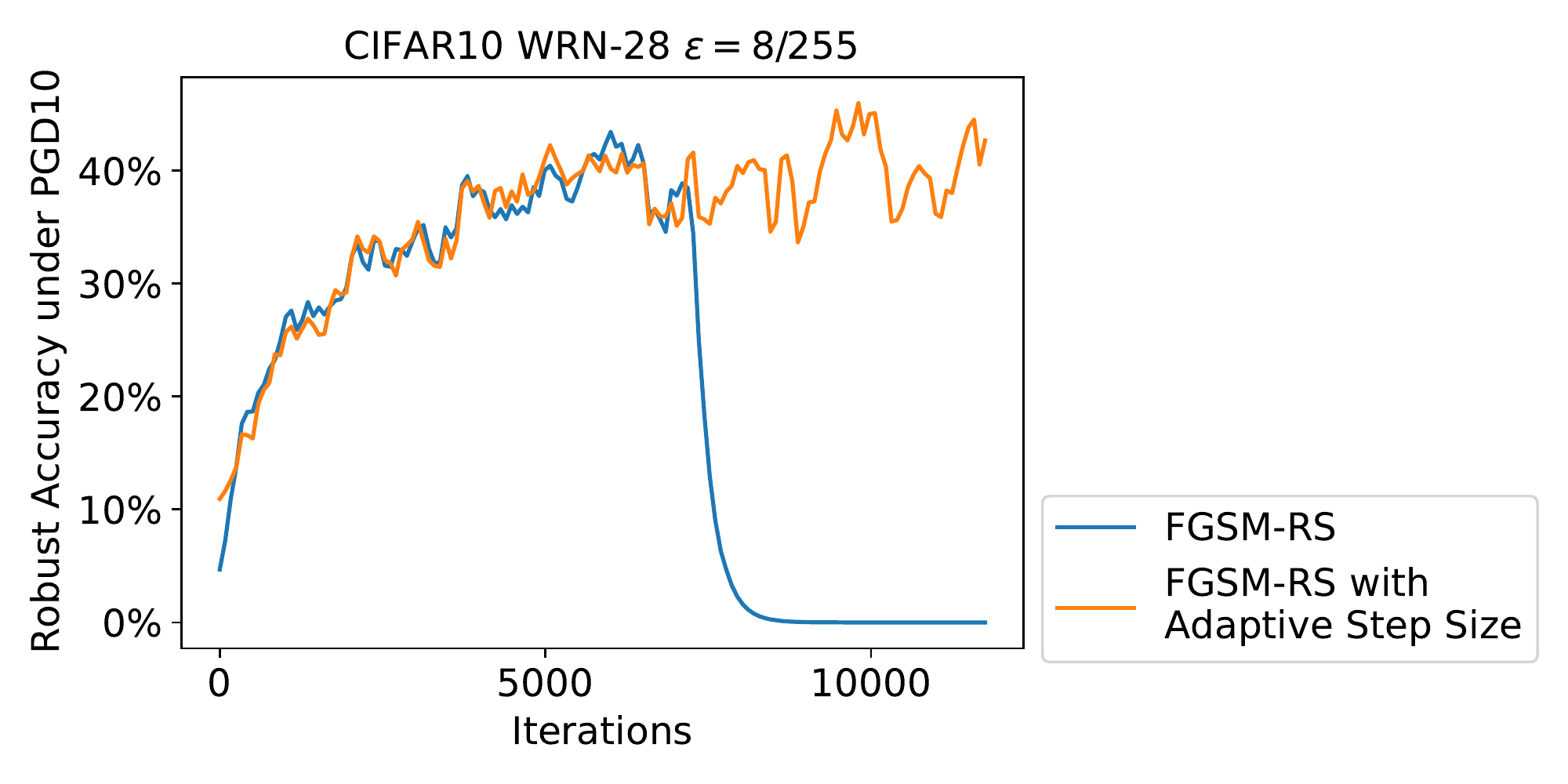}
        \vspace{-0.6cm}
        \caption{}
    \label{fig:adapt_fgsm_rs}
    \end{subfigure}
    \vspace{-0.3cm}
    \caption{(a) The loss gap of training instances between PGD10 and FGSM-RS $\ell(\rvx^{\text{PGD}}, y) - \ell(\rvx^{\text{FGSM-RS}}, y)$ with different step sizes for a FGSM-RS trained robust model; (b) The test robust accuracy of the models trained by FGSM-RS with different step sizes. (c) Accuracy of a WideResNet-28-10 under PGD10 of FGSM-RS and FGSM-RS with adaptive step size. }
    \label{fig:stepsize}
    \vspace{-0.5cm}
\end{figure}

\section{Algorithms}

From our analysis in \Cref{sec:motivation}, the step size of the inner maximization plays an important role for the performance of the single step methods. Overly large step size draws all FGSM-perturbed noise near the boundary, causing catastrophic overfitting and thus the robust accuracy under PGD decreases to zero. However, we cannot simply reduce the step size. As shown in \Cref{fig:loss_gap} and \ref{fig:accuracy_stepsize}, increasing step size can strengthens the adversarial attack and improves the robust accuracy.

To strengthen the attack as much as possible as well as avoid the catastrophic overfitting, we advocate utilizing the instance-wise step-size. The analysis in \Cref{sec:motivation} shows that we should use small step sizes for instances with large gradient norms to prevent catastrophic overfitting, and large step sizes for instances with small gradient norms to the strengthen the attack. 
Thus, we use the moving average of the gradient norm
\begin{equation}
    v_i^j = \beta v_i^{j-1} + (1-\beta) \|\nabla_{\tilde{\rvx}_i} \ell(\tilde{\rvx}_i, y_i; \rvtheta)\|_2^2\;,
\end{equation}
to adjust the step size $\alpha_i^j$ for the $\rvx_i$ at the $j$-th epoch. Here, $\tilde{\rvx}_i$ is the initialization of $\rvx_i$ and $\beta$ is the momentum factor stabilizing the step size. The step size $\alpha_i^j$ is inversely proportional to $v_i^j$:
\begin{equation}
    \alpha_i^j = \gamma/(c + \sqrt{v_i^j})\;,
\end{equation}
where $\gamma$ is a pre-defined learning rate and $c$ is a constant preventing $\alpha_i^j$ from being too large.
We incorporate the adaptive step size $\alpha_i^j$ with FGSM-RS, which randomly initializes the perturbation at the inner maximization step. The results are shown in \Cref{fig:adapt_fgsm_rs}, where the catastrophic overfitting does not occur by adaptive step size.
In addition, the average step size of the adaptive step size method is $10.8/255$, which is even larger than the fixed step size $8/255$ in FGSM-RS, leading to a stronger attack and better adversarial robustness.

Random initialization limits the magnitude of perturbations for instances with small step size, weakening the attack strength. In order to make the whole space within the adversarial budget reachable, we consider the previous initialization in ATTA \cite{atta}, which utilizes the transferability of adversarial examples and uses the adversarial perturbation obtained in the previous epoch as the initialization for the inner maximization. Combined with the previous initialization, {\name} does not need large $\alpha_i^j$ to  reach the whole $\ell_p$ norm ball.
For each instance, we use adaptive step size $\alpha_i^j$ and perform the following inner maximization to obtain the adversarial examples:
\begin{equation}
    \rvx_{i}^{j} = \Pi_{\gB_p(\rvx_i, \varepsilon)}[\rvx_{i}^{j-1} + \alpha_i^j \cdot \text{sgn} (\nabla_{\rvx_{i}^{j-1}}\ell(\rvx_{i}^{j-1}, y_i; \rvtheta))],
    \label{eqn:adaptatta}
\end{equation}
where $\rvx_i^j$ is the adversarial example at the $j$-th epoch.
Then the parameter $\rvtheta$ is updated with $\rvx_{ i}^{j}$
\begin{equation}
    \rvtheta = \rvtheta - \eta \nabla_\rvtheta \ell(\rvx_{i}^{j}, y_i; \rvtheta)\;.
\end{equation}
In contrast to previous methods \cite{fgsmga, kim2020understanding} that needs large computational overhead to resolve the problem of catastrophic overfitting, the overhead of {\name} is negligible, since the input gradient $\nabla_{\rvx_i^{j-1}} \ell(\rvx_i^{j-1}, y_i; \rvtheta)$ is already calculated in the attack step in \Cref{eqn:adaptatta}. Thus, calculating the pre-conditioner $v_i^j$ and the step size $\alpha_i^j$ does not need additional forward-backward passes of the network. The training time of {\name} is almost the same as ATTA \cite{atta} and FGSM-RS \cite{fastat}. The detailed algorithm of {\name} is shown in \Cref{alg:adaptatta}.

{
\begin{algorithm}[t] 
    \caption{\name} 
    \label{alg:adaptatta} 
    \begin{algorithmic}[1]
    \REQUIRE Training set $\gD$, The model $f_\rvtheta$ with loss function $\ell$, Adversarial budget $\varepsilon$
    \ENSURE Optimized model $f_{\rvtheta^*}$\\ 
    \STATE $v_i^0=0$ for $i=1, \cdots, n$
    \STATE $\rvx_i^0$ = $\rvx_i$ + Uniform($-\varepsilon, \varepsilon$) for $i=1, \cdots, n$
    \FOR{$ j = 1$ to $N$}
        \FOR {$\rvx_i, y_i \in \gD$}
            \STATE $v_i^{j} = \beta v_i^{j-1} + (1-\beta) \|\nabla_{\rvx_i^{j-1}} \ell(\rvx_i^{j-1}, y_i; \rvtheta)\|_2^2$
            \STATE $\alpha_i^{j} = \gamma/(c + \sqrt{v_i^j})$
            \STATE $\rvx_i^{j} = \Pi_{\gB_p(\rvx_i, \varepsilon)}[\rvx_i^{j-1} + \alpha_i^{j}\cdot \text{sgn} (\nabla_{\rvx_i^{j-1}}\ell(\rvx_i^{j-1}, y; \rvtheta))]$
            \STATE $\rvtheta = \rvtheta - \eta \nabla_{\rvtheta}\ell(\rvx_i^{j}, y; \rvtheta))$
        \ENDFOR
    \ENDFOR
    \end{algorithmic}
\end{algorithm}
}

\noindent\textbf{Theoretical Analysis of {\name}.}
We analyze the convergence of {\name} with $L_\infty$ adversarial budget. The proof is deferred to \Cref{sec:proof}. 
Given the objective function
\begin{equation}
    \phi(\rvtheta, \rvx) = \frac{1}{n} \sum_{i=1}^n \ell({\rvx}_i, y_i; \rvtheta)\;,
    \label{eqn:phi}
\end{equation}
the minimax problem can be formulated as follows:
\begin{equation}
    \min_\rvtheta \max_{\rvx^*=[{\rvx}_1^*, {\rvx}_2^*, \cdots, {\rvx}_n^*] \in \gB_\infty(\rvx, \varepsilon)} \phi(\rvtheta, \rvx^*)\;,
\end{equation}
where $\rvx^*$ is the optimal adversarial example depending on $\rvtheta$. We consider the minimax optimization in convex-concave and smooth setting. And the loss function $\ell$ satisfies the following assumptions.
\newtheorem{assumption}{Assumption}[section]
\begin{assumption}
The training loss function $\ell$ satisfies the following constraints:

\noindent 1. $\ell$ in convex and $L_\theta$-smooth in $\rvtheta$;
    $\rvtheta$ and the gradient of $\rvtheta$ are bounded in the $L_2$ norm balls
    $$
    \|\rvtheta- \rvtheta^*\|_2 \le D_{\theta, 2}, \quad \frac{1}{n}\sum_{i=1}^n \|\nabla_\rvtheta \ell(\rvx_i', y_i; \rvtheta)\|_2^2, \le G_{\theta, 2}^2\;,
    $$
    where $\rvtheta^* = \argmin_{\rvtheta} \max_{\rvx^* \in \gB_\infty(\rvx, \varepsilon)} \phi(\rvtheta, \rvx^*)$.
    
\noindent 2. $\ell$ in concave and $L_x$-smooth in each $\rvx_i$.
    $\rvx_i \in \mathbb{R}^d$ is bounded in an $L_\infty$ norm ball with $D_{x, \infty} = 2\varepsilon$. For any $\rvx$ and $\rvx'$,
    $
    \|\rvx- \rvx'\|_\infty \le D_{x, \infty}
    $,
    and the gradients of the inputs also satisfy
    $$
    \|\nabla_{\rvx_i'}\ell(\rvx_i', y_i; \rvtheta)\|_2^2 \le G_{x_i, 2}^2, \  \sum_{i=1}^n\|\nabla_{\rvx_i'}\ell(\rvx_i', y_i; \rvtheta)\|_2^2 \le G_{x, 2}^2
    $$
    \vspace{-0.4cm}
\label{asp:convexconcave}
\end{assumption}
We average the trajectory of $T$-steps $\bar{\rvtheta}^T = \frac{\sum_{t=1}^T \rvtheta^t}{T}$ and $\bar{\rvx}^T = \frac{\sum_{t=1}^T\rvx^{t+1}}{T}$
to get the near optimal points. It is a standard technique for analyzing stochastic gradient methods \cite{adagrad}. The convergence gap
$    \max_{\rvx^* \in \gB_\infty(\rvx, \varepsilon)} \phi(\bar{\rvtheta}^T, {\rvx}^*) - \max_{\rvx^* \in \gB_\infty(\rvx, \varepsilon)} \phi(\rvtheta^*, {\rvx}^*)
$
is upper bounded by the regret $R(T)$
\begin{equation} \label{eq:regret}
    R(T) = \sum_{t=1}^T[\max_{\rvx^* \in \gB_\infty(\rvx, \varepsilon)}\phi(\rvtheta^t, \rvx^*)  - \min_{\rvtheta^*}\phi(\rvtheta^*, \rvx^t)]\;.
\end{equation}
\newtheorem{lemma}{Lemma}[section]
\begin{lemma}
For $\ell$ satisfying assumption \ref{asp:convexconcave}, the objective function $\phi$ defined in \Cref{eqn:phi}
$$
\max_{\rvx^* \in \gB_\infty(\rvx, \varepsilon)} \phi(\bar{\rvtheta}^T, {\rvx}^*) - \min_{\rvtheta^*} \max_{\rvx^* \in \gB_\infty(\rvx, \varepsilon)} \phi(\rvtheta^*, {\rvx}^*) \le \frac{R(T)}{T}
$$
\label{theo:regret}
\vspace{-0.5cm}
\end{lemma}

\noindent\textbf{Adaptive Stochastic Gradient Descent Block Coordinate Ascent (ASGDBCA).} {\name} can be formulated as ASGDBCA, which randomly picks an instance $\rvx_k$ at the step $t$, applying stochastic gradient descent to the parameter $\rvtheta$ and adaptive block coordinate ascent to the input ${\rvx}$. Unlike SGDA \cite{sgda}, where all dimensions of $\rvx$ get updated in each iteration, ASGDBCA only updates some dimensions of $\rvx$. ASGDBCA first calculates the pre-conditioner $v_i^t$ as
\begin{equation*}
    \begin{aligned}
        &v_k^{t+1} =         
        \begin{cases}
            \beta v_i^{t} + (1-\beta) \|\nabla_{\rvx_i^t}\ell(\rvx_i^t, y_k; \rvtheta^t)\|_2^2 &  i=k \\
            v^t_i &  i \neq k
        \end{cases}, \qquad
        \hat{v}^{t+1}_i = \text{max}(\hat{v}^{t}_i, v^{t+1}_i)\;.
    \end{aligned}
\end{equation*}
Then $\rvx$, $\rvtheta$ are optimized with 
\begin{equation*}
    \begin{aligned}
        &\rvx^{t+1}_i = 
        \begin{cases}
            \Pi_{\mathcal{B}_\infty(\rvx_i, \varepsilon)}[\rvx^{t}_{i}+\frac{\eta_x}{\sqrt{\hat{v}_{i}^{t+1}}} \nabla_{\rvx_i^t}\ell(\rvx_i^t, y_i; \rvtheta^t)]  &  i=k \\
            \rvx_i^t &  i \neq k
        \end{cases}, \quad
        \rvtheta^{t+1} = \rvtheta^{t}-\eta_\theta \nabla_{\rvtheta}\ell(\rvx_k^{t+1}, y_k; \rvtheta^t)\;.
    \end{aligned}
\label{eqn:ada_sgdbca}
\end{equation*}
The difference between ASGDBCA and {\name} is $\hat{v}_k^t$. To prove the convergence of ASGDBCA, the pre-conditioner needs to be non-decreasing. Otherwise, {\name} may not converge like ADAM \cite{amsgrad}. However, the non-convergent version of ADAM actually works better for neural networks in practice \cite{adam}. Therefore, {\name} still uses $v_k^t$ as the pre-conditioner.

\newtheorem{thm}{Theorem}[section]
\begin{thm}[Regret Bound for ASGDBCA]
Under Assumption \ref{asp:convexconcave}, with $\eta_\theta = \frac{D_{\theta,2}}{G_{\theta,2}\sqrt{T}}$ and $\eta_x = \frac{\sqrt{d}D_{x, \infty}}{\sqrt{T}(1-\beta)^{-1/4}}$, the regret of ASGDBCA is bounded by:
$$
\begin{aligned}
    R^{\text{ASGDBCA}}(T) \le & G_{\theta,2} D_{\theta,2} \sqrt{T} + \frac{D_{x,\infty}\sum_{i=1}^n G_{x_i, 2} \sqrt{dT}}{n(1-\beta)^{1/4}} + \frac{dL_xD_{x,\infty}^2}{2n^2\sqrt{1-\beta}}
\end{aligned}
$$
\vspace{-0.5cm}
\label{thm:asgdbca}
\end{thm}

\noindent\textbf{Comparison with the Non-adaptive Version.} The non-adaptive version of {\name} is ATTA, which can be formulated as the Stochastic Gradient Descent Block Coordinate Ascent (SGDBCA):
\begin{equation*}
    \begin{aligned}
        \rvx^{t+1}_{i} &=  
        \begin{cases}
            \Pi_{\mathcal{B}_\infty(\rvx_i, \varepsilon)}[\rvx^{t}_{i}+\eta_x \nabla_{\rvx_i^t}\ell(\rvx_i^t, y_i; \rvtheta^t)]  &  i=k \\
            \rvx_i^t &  i \neq k
        \end{cases}, \quad
        \rvtheta^{t+1} &= \rvtheta^{t}-\eta_\theta \nabla_{\rvtheta}\ell(\rvx_k^{t+1}, y_k; \rvtheta^t)\;,
    \end{aligned}
\end{equation*}

\begin{thm}[Regret Bound for SGDBCA]
Under assumption \ref{asp:convexconcave}, with constant learning $\eta_\theta = \frac{D_{\theta,2}}{G_{\theta,2}\sqrt{T}}$ and $\eta_x = \frac{\sqrt{nd}D_{x,\infty}}{G_{x,2}\sqrt{T}}$, the regret $R^{\text{SGDBCA}}(T)$ of SGDBCA is bounded by:
$$
R^{\text{SGDBCA}}(T) \le G_{\theta,2} D_{\theta,2} \sqrt{T} + G_{x,2}D_{x,\infty}\sqrt{\frac{dT}{n}} + \frac{d L_x D_{x, \infty}^2}{2n}
$$
\label{thm:sgdbca}
\vspace{-0.5cm}
\end{thm}

Theorem \ref{thm:asgdbca} and \ref{thm:sgdbca} shows that ASGDBCA converges faster than SGDBCA. When $T$ is large, the third term of the regret in both SGDBCA and ASGDBCA is negligible. Consider their first terms are the same, the main difference is the regret bound about $\rvx$ in the second term: $G_{x,2}D_{x,\infty}\sqrt{\frac{dT}{n}}$ and $\frac{D_{x,\infty}\sum_{i=1}^n G_{x_i, 2} \sqrt{dT}}{n(1-\beta)^{1/4}}$.
The ratio between them is 
$$
\text{Ratio} = \frac{1}{(1-\beta)^{\frac{1}{4}}}\sqrt{\frac{\sum_{i=1}^n G_{x_i,2}^2}{n}\Big/(\frac{\sum_{i=1}^{n} G_{x_i,2}}{n})^2}
$$
The Cauchy-Schwarz inequality indicates the ratio is always larger than 1. The gap between ASGDBCA and SGDBCA gets larger when $G_{x_i,2}$ has long-tailed distribution, which demonstrates the relatively faster convergence of {\name} than the non-adaptive counterparts. We show the empirical histogram of $G_{x_i,2}$ of a RN-18 and the ratio in \Cref{fig:theory} in the Appendix, 
which demonstrates the long-tailed distribution for common datasets.

\begin{table*}[t!]
    \caption{Accuracy and training time of different methods on CIFAR10, CIFAR100 and ImageNet. {\name} improves the robust accuracy under various attacks including PGD10, PGD50 and AutoAttack (AA). The method ``\textit{PGD10}" refers to the standard AT using PGD10 for the inner maximization. Note that, we do not have enough computational resources to perform standard AT and SSAT on ImageNet because of computational complexity. Besides, we are unable to train the ResNet-50 on ImageNet with FGSM-GA as its memory requirement exceeds the maximum GPU memory of our devices (\textit{i.e.} NVIDIA Tesla V100). For CIFAR10 and CIFAR100, the training time is evaluated on a single GPU. And we use two GPUs to train the models for ImageNet. We use default step size from the original papers for the baselines so that catastrophic overfitting seldom happens in these methods.}
    \vspace{-0.2cm}
    \begin{subtable}{\linewidth}
        \centering
        \captionsetup{font=normalsize}
        \caption{CIFAR10 with adversarial budget $\varepsilon=8/255$. The accuracy with $\varepsilon=12/255$ and $16/255$ is in the \Cref{tab:cifar10_app}}
        \setlength\tabcolsep{3.4pt}
        \begin{tabular}{l c c c c c c c c c c}
            \hline
            \multirow{2}*{Methods} & \multicolumn{5}{c}{ResNet-18} & \multicolumn{5}{c}{WideResNet-28-10} \\
            \cmidrule(r){2-6}  \cmidrule(r){7-11}
    
            & Clean & PGD10 & PGD50 & AA & Time(h) & Clean & PGD10 & PGD50 & AA & Time(h) \\ \hline
            \textit{PGD10} & \textit{80.13} & \textit{50.59} & \textit{48.94}  & \textit{45.97} & \textit{1.23} & \textit{85.00} & \textit{55.51} & \textit{53.53}  & \textit{51.27} & \textit{8.49}\\         \hline
            FreeAT  & 78.37 & 40.90 &  39.02 &  36.00 & 0.33 & 84.54 & 46.09 &  43.80 &  41.19 & 2.31 \\
            YOPO  & 74.72 & 37.51 & 35.79 & 33.21 & 0.28 & 82.92 & 44.62 & 42.14 & 40.23 & 1.90 \\
            FGSM-RS& \textbf{83.99} & 48.99 & 46.36 & 42.95 & \textbf{0.22} & 80.21 & 0.01 & 0.00 & 0.00 & 1.67 \\
            FGSM-GA& 80.10 & 49.14 & 47.21 & 43.44 & 0.57 & 75.84 & 45.57 & 43.28 & 39.44 & 3.82 \\
            SSAT & 88.83 & 42.31 & 38.99 & 37.06 & 0.61 & 90.40 & 44.04 & 40.40 & 38.82 & 3.53\\
            ATTA & 82.16 & 47.47 & 45.32 & 42.51 & 0.30 & 85.90 & 51.52 & 48.94 & 46.84 & 1.70\\
            {\name} & 81.22 & \textbf{50.03} & \textbf{48.18} & \textbf{45.38} & 0.30 & \textbf{85.96} & \textbf{53.43} & \textbf{51.03} & \textbf{48.72} & \textbf{1.63} \\ 
            \hline 
        \end{tabular}

        \label{tab:cifar10}
        \vspace{0.2cm}
    \end{subtable}
    
    \begin{subtable}{\linewidth}
        \centering
        \captionsetup{font=normalsize}
        \caption{CIFAR100 with adversarial budget $\varepsilon=8/255$. The accuracy with $\varepsilon=4/255$ and $12/255$ is in the \Cref{tab:cifar100_app}
        }
        \setlength\tabcolsep{3.4pt}
        \begin{tabular}{l c c c c c c c c c c}
            \hline
            \multirow{2}*{Methods} & \multicolumn{5}{c}{ResNet-18} & \multicolumn{5}{c}{WideResNet-28-10} \\
            \cmidrule(r){2-6}  \cmidrule(r){7-11}
            & Clean & PGD10 & PGD50 & AA & Time(h) & Clean & PGD10 & PGD50 & AA & Time(h) \\ \hline
            \textit{PGD10}&  \textit{54.08} & \textit{28.03} & \textit{27.23}  & \textit{23.04}  & \textit{1.32} & \textit{60.04} & \textit{31.70} & \textit{30.67}  & \textit{27.11} & \textit{8.53} \\
            \hline 
            FreeAT  & 50.56 & 19.57 &  18.58 &  15.09 & 0.33 & 59.38 & 24.41 & 23.00 & 19.60 & 2.30 \\
            YOPO & 51.55 & 20.65 & 19.17 & 16.05 & 0.29 & 50.35 & 19.44 & 18.36 & 15.43 & 1.92 \\
            FGSM-RS& \textbf{59.35} & 26.40 & 24.29 & 19.73 & \textbf{0.21} & 51.83 & 0.00 & 0.00 & 0.00 & \textbf{1.60} \\
            FGSM-GA& 50.61 & 24.48 & 24.07 & 19.42 & 0.57 & 54.29 & 25.86 & 24.56 & 20.74 & 3.80   \\
            SSAT & 71.03 & 9.79 & 4.80 & 1.09 & 0.62 & 75.01 & 0.21 & 0.01 & 0.00 & 3.50\\
            ATTA & 57.21 & 25.76 & 24.90 & 21.03 & 0.28 & \textbf{63.04} & 28.93 & 27.18 & 24.42 & 1.63\\
            {\name} &  55.49 & \textbf{27.68} & \textbf{26.60} & \textbf{22.62} & 0.31 & 62.34 & \textbf{29.89} & \textbf{28.35} & \textbf{25.03} & 1.61  \\ \hline
        \end{tabular}

        \vspace{0.2cm}

        \label{tab:cifar100}
    \end{subtable}
    
    \begin{subtable}{\linewidth}
        \centering
        \captionsetup{font=normalsize}

        \caption{ImageNet  with adversarial budget  $\varepsilon=2/255$. The accuracy with $\varepsilon=4/255$ is available in the \Cref{tab:imagenet_app}
        }
        \setlength\tabcolsep{3.4pt}
        \begin{tabular}{l c c c c c c c c c c}
            \hline
            \multirow{2}*{Methods} & \multicolumn{5}{c}{ResNet-18} & \multicolumn{5}{c}{ResNet-50} \\
            \cmidrule(r){2-6}  \cmidrule(r){7-11}
            & Clean & PGD10 & PGD50 & AA & Time(h) & Clean & PGD10 & PGD50 & AA & Time(h) \\ \hline
            FreeAT & 58.80 & 35.56 &  34.78 &  31.77 & \textbf{40.01} & 65.81 & 44.12 & 43.34 & 40.80 & \textbf{108.3} \\
            YOPO  & 47.69 & 28.50 & 28.10 & 25.22 & 48.22 & 55.68 & 33.46 & 32.19 & 29.56 & 111.8 \\
            FGSM-RS& 55.26 & 37.33 & 36.98 & 33.28 & 43.46 & 67.83 & 46.12 & 45.56 & 43.58 & 115.0 \\
            FGSM-GA& 37.01 & 24.15 & 24.05 & 19.98 & 182.7 & / & / & / & / & /   \\
            ATTA & 58.32 & 39.62 & 38.32 & 36.08 & 45.83 & 66.62 & 48.27 & 47.65 & 45.00 & 111.7 \\
            {\name} &  \textbf{61.20} & \textbf{40.84} & \textbf{39.86} & \textbf{37.25} & 45.70 & \textbf{69.10} & \textbf{49.05} & \textbf{48.05} & \textbf{46.01} & 120.4\\ \hline
        \end{tabular}

        \label{tab:imagenet}
    \end{subtable}
    \label{tab:accuracy}
    \vspace{-0.6cm}
\end{table*}

\section{Experiments}

\noindent\textbf{Baselines.} We compare {\name} with the SOTA fast AT algorithms including FreeAT \cite{freeat}, YOPO \cite{yopo}, FGSM-RS \cite{fastat}, FGSM-GA \cite{fgsmga}, SSAT \cite{kim2020understanding} and ATTA \cite{atta}. We also compare {\name} with standard AT whose inner maximization is solved by PGD10, providing a reference for the ideal performance. 

\noindent\textbf{Attack Methods.} We consider three attacks: PGD10, PGD50 \cite{at} and AutoAttack (AA) \cite{autoattack}. Square Attack, a black-box attack, is included in AutoAttack to eliminate the effect of  gradient masking.

\noindent\textbf{Experimental Settings.} {\name} uses the techniques proposed in ATTA \cite{atta}: the adversarial perturbations are transformed according the data augmentation and get reset every several epochs. And the previous initialization is stored in the GPU memory, brings negligible storing latency to {\name}. We consider adversarial attacks with the $\ell_\infty$-norm budget. We evaluate fast AT algorithms on CIFAR10 and CIFAR100 \cite{cifar10} with WideResNet-28-10 (WRN-28-10) \cite{wrn} and ResNet-18 (RN-18), and on ImageNet \cite{deng2009imagenet} with ResNet-18 (RN-18) and ResNet-50 (RN-50). 
While early stopping is widely used in the standard AT \cite{overfitting}, the computational overhead to perform PGD attack on a separate validation set is large. 
Besides, considering the small budget of training time in fast AT, even if early stopping is applied to terminate the training before catastrophic overfitting occurs, the training is far from convergence, resulting in poor performance \cite{fgsmga}. Therefore, we follow the previous works \cite{fgsmga, fastat, atta} and do not use early stopping. We set $\beta=0.5$ and $\gamma/c = 16/255$, which is close to the adversarial budget. And we set $c=0.01$ for CIFAR10 and CIFAR100 and $c=0.1$ for ImageNet.  More detailed experiment settings are in \Cref{sec:expsetting}. Additional experiments are available in the \Cref{sec:addexp}.  
å
\begin{figure*}[t]
    \centering
    \includegraphics[width=\linewidth]{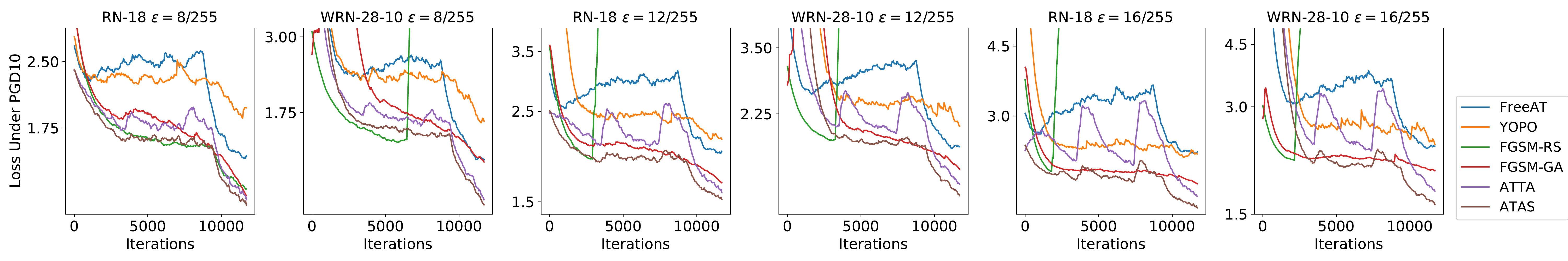}
    \vspace{-0.6cm}
    \caption{Robust training cross-entropy loss under PGD10 of CIFAR10 with different network architectures and adversarial budgets. The curve is smoothed to clearly show the convergence. }
    \label{fig:convergence}
    \vspace{-0.5cm}
\end{figure*}

\begin{figure*}[t]
    \centering
    \includegraphics[width=\linewidth]{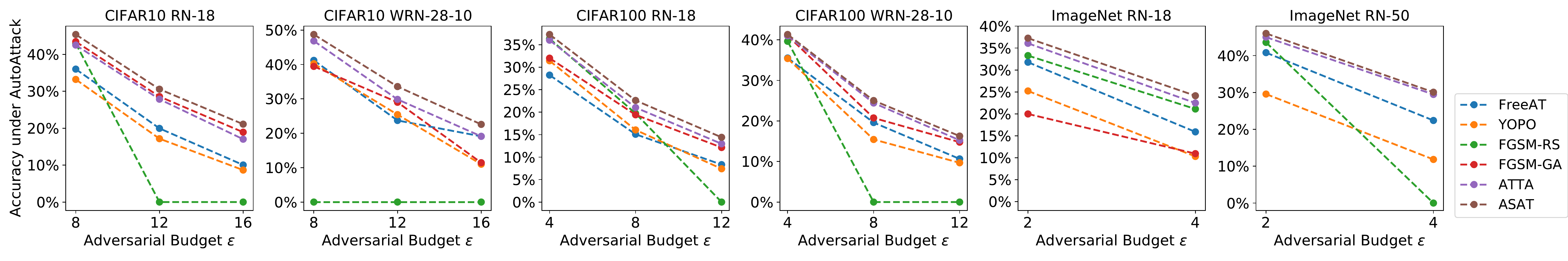}
    \vspace{-0.6cm}
    \caption{Robust accuracy under AutoAttack for different datasets on different network architectures with varying adversarial budgets. {\name} achieves the highest robust accuracy in these cases. The accuracy numbers can be found in the \Cref{sec:addexp_acc}.}
    \label{fig:acc_curve}
    \vspace{-0.6cm}
\end{figure*}

\begin{table}[!t]
\vspace{0.3cm}
    \centering
    \caption{Ablation study of hyperparameters $\gamma$ (left) and $c$ (right) on CIFAR10 and RN-18 under AA.}
    \begin{subtable}{0.51\linewidth}
        \captionsetup{font=normalsize}
        \setlength\tabcolsep{2.0pt}
        \begin{tabular}{c c c c c c c } \hline
            $\gamma/0.01*255$ & 12 & 14 & 16 & 18 & 20 \\ \hline
            $\varepsilon=8/255$ & 45.20 & 45.21 & 45.38 & 45.50 & 45.60 \\
            $\varepsilon=12/255$ & 30.84 & 31.06 & 30.56 & 31.21 & 31.04 \\
            $\varepsilon=16/255$ & 21.38 & 21.23 & 21.09 & 21.13 & 20.94 \\ \hline
        \end{tabular}
    \end{subtable}
    \begin{subtable}{0.48\linewidth}
        \captionsetup{font=normalsize}
        \setlength\tabcolsep{2.0pt}
        \begin{tabular}{c c c c c c} \hline
            $c$ & 0.005 & 0.007 & 0.01 & 0.02 & 0.04 \\ \hline
            $\varepsilon=8/255$ & 45.01 & 45.28 & 45.38 & 45.52 & 45.48 \\
            $\varepsilon=12/255$  & 30.08 & 30.80 & 30.56 & 30.69 & 30.52  \\
            $\varepsilon=16/255$ & 20.36 & 20.84 & 21.09 & 21.07 & 20.48 \\ \hline
        \end{tabular}
    \end{subtable}
    \label{tab:ablation}
    \vspace{-0.7cm}
\end{table}

\noindent\textbf{Convergence.}
\Cref{fig:convergence} shows the curve of the training loss
$
    \max_{\rvx^*=[{\rvx}_1^*, \cdots, {\rvx}_n^*] \in \gB_\infty(\rvx, \varepsilon)} \phi(\rvtheta, \rvx^*)
$
on CIFAR10 with different network architectures and different adversarial budgets, where $\rvx^*$ is approximated by PGD10 and the objective function $\phi$ is approximated by mini-batches of training instances at each step. {\name} achieves smaller robust training loss at the end of training, demonstrating the faster convergence of {\name} than ATTA and other baselines. We also show the relationship between gradient norm distribution and convergence gap between ATTA and {\name} in \Cref{sec:addexp_convergence}.

\noindent\textbf{Robust Accuracy.} We provide our main results in \Cref{tab:accuracy}, showing the robust accuracy of CIFAR10, CIFAR100 and ImageNet, respectively. \Cref{fig:acc_curve} shows the robust accuracy under AutoAttack for different adversarial budgets, whose numbers are provided in the \Cref{sec:addexp_acc}.

\noindent\textbf{CIFAR10 and CIFAR100.} As shown in \Cref{tab:cifar10}, The robust accuracy of FreeAT and YOPO is much lower than the other methods. While FGSM-RS maintains non-trivial robust accuracy when using RN-18, it suffers from catastrophic overfitting when using large networks such as WRN-28-10. The regularizer in FGSM-GA prevents catastrophic overfitting. However, it may over-regularize the network so that the clean accuracy and the robust accuracy decrease on WRN-28-10. In addition, the regularizer also brings computational overhead: FGSM-GA needs nearly double training time compared with other methods. {\name} achieves the best robust accuracy among all fast AT algorithms while keeping the training time nearly the same. Furthermore, for small networks like RN-18, the performance of {\name} is on par with standard AT (PGD10) but needs only one fifth of the training time.
\Cref{tab:cifar100} shows the robust accuracy on CIFAR100 and {\name} also outperforms other algorithms. Catastrophic overfitting also happens in SSAT even if the losses of inner points are checked.

\noindent\textbf{ImageNet.} 
ATTA and {\name} need to memorize the adversarial noise for the whole training set. Since frequently loading and storing from the disks significantly lowers the training speed, all perturbations should be stored in the memory. Thus, we utilize the local property of the adversarial examples \cite{huang2020corrattack} and only store the interpolated perturbation in the memory. We resize the perturbations from $224\times 224$ to $32\times 32$ for storage and up-sample it back when used as the initialization for the next epoch. The detailed algorithm is deferred to the \Cref{sec:atas_imagenet}.
\Cref{tab:imagenet} shows the robust accuracy on ImageNet on $\varepsilon=2/255$. {\name} still has higher robust accuracy than all baselines. FGSM-GA needs the calculate the second order gradient of the parameters, which needs huge amount of GPU memory. Thus, we could not train a big network such as ResNet-50 on ImageNet. 

\noindent\textbf{Robust accuracy at different adversarial budgets.} \Cref{fig:acc_curve} shows the robust accuracy of fast AT algorithms under AutoAttack on different datasets, network architectures and adversarial budgets.
The robust accuracy decreases when enlarging the adversarial budget, but {\name} always outperforms all the baselines for different adversarial budgets, datasets and network architectures. This demonstrates that the improvement of {\name} is consistent. 

\noindent\textbf{Ablation Study.} \Cref{tab:ablation} provides the ablation study on hyperparameters, showing that {\name} is not sensitive to them.  Besides, as the only different between {\name} and ATTA is the step size, the superior performance of {\name} over ATTA forms a ablation study to demonstrate the effectiveness of the adaptive step size. The changes of gradient norm and step size of {\name} is shown in \Cref{sec:addexp_step}.

\section{Conclusion}

In this paper, we investigate catastrophic overfitting from the perspective of training instances and show that instances with large gradient norms are more likely to cause catastrophic overfitting in the single-step fast AT methods. 
This finding motivates the adaptive training method, {\name}, which applies the adaptive step size of inner maximization inversely proportional to the input gradient norm.
We theoretically analyze the convergence of {\name}, showing that our method converges faster than the non-adaptive counterpart especially when the distribution of input gradient norm is long-tailed.
Extensive experiments on CIFAR10, CIFAR100 and ImageNet with different network architectures and adversarial budgets show that {\name} mitigates catastrophic overfitting and achieves higher robust accuracy under various strong attacks.  

\bibliography{ref}
\bibliographystyle{plain}

\newpage

\appendix

In the Appendix, we provide additional materials to supplement our main submission. 
In \Cref{sec:proof}, we provide the proof of \Cref{theo:regret}, Theorem \ref{thm:asgdbca} and Theorem \ref{thm:sgdbca} in the main text.
In \Cref{sec:addexp}, we report additional experimental results including the catastrophic overfitting in ATTA and the robust accuracy of various adversarial budgets.
In \Cref{sec:expsetting}, we provide the detailed hyperparameters and experimental settings.

\section{Proof of Section 4.1}
\label{sec:proof}
We provide the proof of the regret bound in this section.

\subsection{Proof the Lemma 4.1}

\begin{proof}
\begin{equation}
    \begin{aligned}
    &\max_{\rvx^* \in \gB_\infty(\rvx, \varepsilon)}\phi(\bar{\rvtheta}^T, {\rvx}^*) - \min_{\rvtheta^*} \max_{\rvx^* \in \gB_\infty(\rvx, \varepsilon)} \phi(\rvtheta^*, {\rvx}^*)  \\
    \le & \max_{\rvx^* \in \gB_\infty(\rvx, \varepsilon)}\phi(\bar{\rvtheta}^T, {\rvx}^*) - \min_{\rvtheta^*}\phi(\rvtheta^*, \bar{\rvx}^T) \\
    = & \max_{\rvx^* \in \gB_\infty(\rvx, \varepsilon)}\phi(\frac{\sum_{t=1}^T \rvtheta^t}{T}, {\rvx}^*) - \min_{\rvtheta^*}\phi(\rvtheta^*, \frac{\sum_{t=1}^T\rvx^{t+1}}{T}) \\
    \le & \frac{\min_{\rvtheta^*}\sum_{t=1}^T\phi(\rvtheta^t, \rvx^*) - \min_{\rvtheta^*}\sum_{t=1}^T\phi(\rvtheta^*, \rvx^{t+1})}{T} \\
    \le & \frac{\sum_{t=1}^T(\max_{\rvx^* \in \gB_\infty(\rvx, \varepsilon)}\phi(\rvtheta^t, \rvx^*) - \min_{\rvtheta^*}\phi(\rvtheta^*, \rvx^{t+1}))}{T} \\
    = & \frac{R(T)}{T}.
    \end{aligned}
\end{equation}

The first and the third inequality follows the optimality condition and the second inequality uses
the Jensen inequality.
\end{proof}
Before moving to the proof of Theorem 4.1 and 4.2, we define several notations of gradients as follows:
\begin{equation}
\begin{aligned}
    \hat{g}_\theta^k(\rvtheta, \rvx) &\equiv \nabla_{\rvtheta} \ell(\rvx_k, y_k; \rvtheta), \\
    g_\theta(\rvtheta, \rvx) &\equiv \mathbb{E}_k \hat{g}_\theta^k(\rvtheta, \rvx) = \nabla_{\rvtheta} \phi(\rvtheta, \rvx), \\
    g_x^k(\rvtheta, \rvx) & \equiv -\nabla_{\rvx_k} \ell(\rvx_k, y_k; \rvtheta), \\
     g_x(\rvtheta,\rvx) & \equiv [g_x^1(\rvtheta, \rvx), \cdots, g_x^n(\rvtheta, \rvx)] = -n\nabla_{\rvx} \phi(\rvtheta, \rvx).
\end{aligned}
\end{equation}
And $\mathbb{E}_k$ means the average over $k$.
\subsection{Proof of Theorem 4.2 (SGDBCA)}
\begin{proof}
Let $\eta_x = h_x \eta_\theta$. At step $t$, SGDBCA picks a random instance indexed by $k$ from $\{1, \cdots, n\}$ and updates its adversarial perturbation. Then we have the following inequality:
\begin{align}
    \|\rvx^{t+1}_k - \rvx^*\|_2^2 &= \|\Pi_{\gB_\infty(\rvx, \varepsilon)}(\rvx^{t}_{k}-h_x\eta_\theta^{t} g_x^k(\rvtheta^t, \rvx^{t})) - \rvx^*\|_2^2 \nonumber\\ &\le \|\rvx^{t}_{k}-h_x\eta_\theta^{t} g_x^k(\rvtheta^t, \rvx^{t}) - \rvx^*\|_2^2.
\end{align}
Hence
\begin{align}
    \left\|\rvx^{t+1}_k - \rvx^*\right\|_2^2 \le& \left\|\rvx^t_{k}-\rvx_{k}^*\right\|_2^2 + (h_x \eta_\theta)^2 \left\|g_x^k(\rvtheta^t, \rvx^{t})\right\|_2^2 -  2h_x\eta_\theta g_x^k(\rvtheta^t, \rvx^{t})^{\top}(\rvx^t_{k} - \rvx_{k}^*).
\end{align}
Rearranging the inequality, it is easy to get:
\begin{align}
    2\eta_\theta g_x^k(\rvtheta^t, \rvx^{t})^{\top}(\rvx^t_{k} - \rvx_{k}^*) \le & \frac{\left\|\rvx^t_{k}-\rvx_{k}^*\right\|_2^2 - \left\|\rvx^{t+1}_{k} -\rvx_{k}^* \right\|_2^2}{h_x} + 
     h_x(\eta_\theta)^2 \left\|g_x^k(\rvtheta^t, \rvx^t)\right\|_2^2. \label{eqn:x_oracle}
\end{align}
Similarly, we have:
\begin{align}
    2\eta_\theta g_x^k(\rvtheta^t, \rvx^{t+1})^{\top}(\rvtheta^t - \rvtheta^*) \le & \left\|\rvtheta^t-\rvtheta^*\right\|_2^2 - \left\|\rvtheta^{t+1} -\rvtheta^* \right\|_2^2 +  (\eta_\theta)^2 \left\|\hat{g}_\theta^k(\rvtheta^t, \rvx^{t+1})\right\|_2^2. \label{eqn:t_oracle} 
\end{align}
Taking expectation over $k$ on the left hand side of \Cref{eqn:x_oracle} and \Cref{eqn:t_oracle}, we get:
\begin{equation}
    \begin{aligned}
        &\mathbb{E}_k \left[g_x^k(\rvtheta^t, \rvx^{t})^{\top}(\rvx^t_{k} - \rvx_{k}^*)\right] = \frac{g_x(\rvtheta^t, \rvx^t)^{\top}(\rvx^t-\rvx^*)}{n},    \\
        &\mathbb{E}_k \left[ \hat{g}^k_\theta(\rvtheta^t, \rvx^{t+1})^{\top}(\rvtheta^t - \rvtheta^*) \right] = g_\theta(\rvtheta^t, \rvx^{t+1})^{\top}(\rvtheta^t - \rvtheta^*). \\
    \end{aligned}
    \label{eqn:in1}
\end{equation}
Taking expectation over $k$ on the right hand side of \Cref{eqn:x_oracle}, we have:
\begin{equation}
\begin{aligned}
    & \mathbb{E}_k \left[\left\|\rvx^t_k-\rvx_{k}^*\right\|_2^2 - \left\|\rvx^{t+1}_{k} -\rvx_{k}^* \right\|_2^2 \right] = \frac{1}{n}\left[\left\|\rvx^t-\rvx^*\right\|_{2}^2 - \left\|\rvx^{t+1} -\rvx^* \right\|_{2}^2 \right], 
\end{aligned}
    \label{eqn:in2}
\end{equation}
and 
\begin{equation}
    \mathbb{E}_k\left[\left\|g_x^k(\rvtheta^t, \rvx^t)\right\|_2^2 \right] = \frac{1}{n} \left\|g_x(\rvtheta^t, \rvx^t)\right\|_{2}^2.
    \label{eqn:in3}
\end{equation}
Considering the convex and concave condition of $\rvx$ and $\rvtheta$
\begin{equation}
\begin{aligned}
    \nabla_{\rvtheta^t} \phi(\rvtheta^t, \rvx^{t+1})^{\top}\left(\rvtheta^*-\rvtheta^t\right) &\leq \phi\left(\rvtheta^*, \rvx^{t+1}\right)-\phi(\rvtheta^t, \rvx^{t+1}), \\
    \phi\left(\rvtheta^t, \rvx^*\right)-\phi(\rvtheta^t, \rvx^t)+\nabla_{\rvx^t} &\leq \phi(\rvtheta^t, \rvx^t)^{\top}\left(\rvx^*-\rvx^{t}\right),
\end{aligned}
\end{equation}
we get:
\begin{equation}
\begin{aligned}
    & g_\theta\left(\rvtheta^{t}, \rvx^{t+1}\right)^{\top}\left(\rvtheta^{t}-\rvtheta^*\right) + \frac{1}{n}g_x\left(\rvtheta^{t}, \rvx^{t}\right)^{\top}\left(\rvx^{t}-\rvx^*\right) \\
    \geq  &\phi\left(\rvtheta^{t}, \rvx^*\right)-\phi\left(\rvtheta, \rvx^{t+1} \right) + \phi\left(\rvtheta^t, \rvx^{t+1}\right) - \phi\left(\rvtheta^t, \rvx^{t}\right).
\end{aligned}
    \label{eqn:in4}
\end{equation}
Combining Eqn \eqref{eqn:x_oracle} to \eqref{eqn:in4}, we obtain the following inequality:
\begin{align}
    2\eta_\theta\left(\phi\left(\rvtheta^{t}, \rvx^*\right)-\phi\left(\rvtheta^*, \rvx^{t+1}\right)\right) \nonumber
    \le  \mathbb{E}_k\Bigg[ & \left\|\rvtheta^t-\rvtheta^*\right\|_2^2 - \left\|\rvtheta^{t+1} -\rvtheta^* \right\|_2^2 + \nonumber
    (\eta_\theta)^2 \left\|g_\theta^k(\rvtheta^t, \rvx^{t+1})\right\|_2^2 + \nonumber\\
    & \frac{1}{nh_x}\left(\left\|\rvx^t-\rvx^*\right\|_{2}^2 - \left\|\rvx^{t+1} -\rvx^* \right\|_{2}^2\right)  + \\
    &\frac{1}{n}h_x(\eta_\theta)^2\left\|g_x(\rvtheta^t, x^t)\right\|_{2}^2 +  \\
    & 2\eta_\theta \left( \phi\left(\rvtheta^t, \rvx^t\right) - \phi\left(\rvtheta^t, \rvx^{t+1}\right)\right)\Bigg]. \nonumber
\end{align}
Considering the update of $\rvx$, we have:
\begin{align}
    & \mathbb{E}_k\left[\phi\left(\rvtheta^t, \rvx^t\right) - \phi\left(\rvtheta^t, \rvx^{t+1}\right)\right] \nonumber\\
    \leq &\mathbb{E}_k\left[\frac{L_x}{2n}\left\|\rvx^t_{k}-\rvx^{t+1}_k\right\|_{2}^{2} + \frac{1}{n}g_x^k(\rvtheta^t, \rvx^t)^\top(\rvx^{t+1}_k-\rvx^t_k)\right] \nonumber\\
    = &\mathbb{E}_k\left[\frac{L_x(h_x\eta_\theta)^2}{2n}\left\|g_x^k(\rvtheta^t, \rvx^{t})\right\|_2^2 - \frac{h_x\eta_\theta}{n} \left\|g_x(\rvtheta^t, \rvx^{t})\right\|_2^2\right] \nonumber\\
    = &\frac{L_x(h_x\eta_\theta)^2}{2n^2}\left\|g_x(\rvtheta^t, \rvx^{t})\right\|_{2}^2 - \frac{h_x\eta_\theta}{n^2}\left\|g_x(\rvtheta^t, \rvx^{t})\right\|_{2}^2.
\end{align}
The above inequality can be rearranged as:
\begin{align}
    2\eta_\theta\left(\phi\left(\rvtheta^{t}, \rvx^*\right)-\phi\left(\rvtheta^*, \rvx^{t+1}\right)\right) 
    \le \mathbb{E}_k\Bigg[ & \left\|\rvtheta^t-\rvtheta^*\right\|_2^2 - \left\|\rvtheta^{t+1} -\rvtheta^* \right\|_2^2 + 
    (\eta_\theta)^2 \left\|g_\theta^k(\rvtheta^t, \rvx^{t+1})\right\|_2^2 + \nonumber\\
    & \frac{1}{nh_x}\left(\left\|\rvx^t-\rvx^*\right\|_{2}^2 - \left\|\rvx^{t+1} -\rvx^* \right\|_{2}^2\right)  + \\ &\frac{n-2}{n^2}h_x(\eta_\theta)^2\left\|g_x(\rvtheta^t, \rvx^t)\right\|_{2}^2 + \\
    &\frac{L_x(h_x)^2(\eta_\theta)^3}{n^2}\left\|g_x(\rvtheta^t, \rvx^{t})\right\|_{2}^2 \Bigg]. \nonumber
\end{align}
Divide both side by $\eta_\theta$, then
\begin{equation}
\begin{aligned}
    2\left(\phi\left(\rvtheta^{t}, \rvx^*\right)-\phi\left(\rvtheta^*, \rvx^{t+1}\right)\right)
    \le \mathbb{E}_k\Bigg[ & \frac{1}{\eta_\theta}\left(\left\|\rvtheta^t-\rvtheta^*\right\|_2^2 - \left\|\rvtheta^{t+1} -\rvtheta^* \right\|_2^2 \right) + 
    \eta_\theta \left\|g_\theta(\rvtheta^t, \rvx^{t+1})\right\|_2^2  + \\
    & \frac{1}{nh_x\eta_\theta}\left(\left\|\rvx^t-\rvx^*\right\|_{2}^2 - \left\|\rvx^{t+1} -\rvx^* \right\|_{2}^2\right)  + \\
    & \frac{n-2}{n^2}h_x\eta_\theta\left\|g_x(\rvtheta^t, \rvx^t)\right\|_{2}^2 + \frac{L_x(h_x\eta_\theta)^2}{n^2}\left\|g_x(\rvtheta^t, x^{t})\right\|_{2}^2 \Bigg].
\end{aligned}
\end{equation}
Summing the bound over $t$, the regret is bounded by:
\begin{equation}
\begin{aligned}
    R(T) \le &  \frac{1}{2\eta_\theta}\left\|\rvtheta^1-\rvtheta^*\right\|_2^2 +\frac{1}{2n\eta_x} \left\|\rvx^1 - \rvx^*\right\|_2^2  + \frac{1}{2}\mathbb{E} \sum_{t=1}^T  \eta_\theta \left\|g_\theta^k(\rvtheta^t, \rvx^{t+1})\right\|_2^2 + \\
    &\frac{1}{2}\mathbb{E} \sum_{t=1}^T  \frac{[(n-2)\eta_x + L_x\eta_x^2]}{n^2} \left\|g_x(\rvtheta^t, \rvx^t)\right\|_2^2.
\end{aligned}
\end{equation}
With the bound of $\rvx$, $\rvtheta$ and their gradients, we can simplify the bound as:
\begin{equation}
\begin{aligned}
    R(T) \le & \frac{D_{\theta,2}^2}{2\eta_\theta} +\frac{dD_{x, \infty}^2}{2\eta_x} + \frac{T\eta_\theta G_{\theta, 2}^2}{2} + \frac{T\eta_x G_{x, 2}^2}{2n} +  \frac{TL_x\eta_x^2G_{x,2}^2}{2n^2}.
\end{aligned}
\end{equation}
Using the inequality of arithmetic and geometric means, the optimal choice is $\eta_\theta = \frac{D_{\theta,2}}{G_{\theta,2}\sqrt{T}}$ and $\eta_x = \frac{\sqrt{nd}D_{x,\infty}}{G_{x,2}\sqrt{T}}$, then
\begin{equation}
R(T) \le G_{\theta,2} D_{\theta,2} \sqrt{T} + G_{x,2}D_{x,\infty}\sqrt{\frac{dT}{n}} + \frac{d L_x D_{x, \infty}^2}{2n}.
\end{equation}
\end{proof}

\subsection{Proof of Theorem 4.1 (ASGDBCA)}
\begin{proof}

Let $\eta_x = h_x \eta_\theta$. At step $t$, ASGDBCA picks a random instance indexed by $k$ from $\{1, \cdots, n\}$. Then
\begin{equation}
\begin{aligned}
    2g_x^k(\rvtheta^t, \rvx^{t})^{\top}(\rvx^t_{k} - \rvx_k^*) \le & \frac{\left\|\rvx^t_k-\rvx_k^*\right\|_2^2}{\eta_\theta h_x}\sqrt{\hat{v}^{t+1}_k} -  \frac{\left\|\rvx^{t+1}_{k} -\rvx_k^* \right\|_2^2}{\eta_\theta h_x}\sqrt{\hat{v}^{t+1}_k} + h_x\eta_\theta \frac{\left\|g_x^k(\rvtheta^t, \rvx^t)\right\|_2^2}{\sqrt{\hat{v}_k^{t+1}}}, \\
    2 g_\theta(\rvtheta^t, \rvx^{t+1})^{\top}(\rvtheta^t - \rvtheta^*) \le &  \frac{\left\|\rvtheta^t-\rvtheta^*\right\|_2^2 - \left\|\rvtheta^{t+1} -\rvtheta^* \right\|_2^2}{\eta_\theta} + 
    \eta_\theta \left\|g_\theta^k(\rvtheta^t, \rvx^{t+1})\right\|_2^2. \\  
\end{aligned}
\end{equation}
Let 
$$
    \hat{V}^t = diag([\underbrace{\hat{v}_1^t, \cdots, \hat{v}_1^t}_{d}, \underbrace{\hat{v}_2^t, \cdots, \hat{v}_2^t}_d, \cdots, \underbrace{\hat{v}_n^t, \cdots, \hat{v}_n^t}_d]),
$$
denote the pre-conditioner of all the coordinates of $\rvx$. Take the expectation over $k$ on the right hand side, then
\begin{equation}
\begin{aligned}
    & \mathbb{E}_k \left[\frac{\left\|\rvx^t_k-\rvx_{k}^*\right\|_2^2 - \left\|\rvx^{t+1}_{k} -\rvx_{k}^* \right\|_2^2}{h_x}\sqrt{\hat{v}^{t+1}_k} \right]
    =  \frac{1}{n}\left[\left\|\rvx^t-\rvx\right\|_{\frac{\sqrt{\hat{V}^{t+1}}}{h_x}}^2 - \left\|\rvx^{t+1} -\rvx \right\|_{\frac{\sqrt{\hat{V}^{t+1}}}{h_x}}^2 \right],     \\
\end{aligned}
\end{equation}
and
\begin{equation}
\begin{aligned}
    \mathbb{E}_k\left[h_x\eta_\theta^2 \frac{\left\|g_x^k(\rvtheta^t, \rvx^t)\right\|_2^2}{\sqrt{\hat{v}_k^{t+1}}} \right] & =  \frac{\eta_\theta^2}{n} \left\|g_x(\rvtheta^t, \rvx^t)\right\|_{\frac{h_x}{\sqrt{\hat{V}^{t+1}}}}^2.
\end{aligned}
\end{equation}
Similar to the proof of SGDBCA, we have:
\begin{equation}
\begin{aligned}
    2\left(\phi\left(\rvtheta^{t}, \rvx^*\right)-\phi\left(\rvtheta^*, \rvx^{t+1}\right)\right)
    \le \mathbb{E}_k\Bigg[ & \frac{1}{\eta_\theta}\left(\left\|\rvtheta^t-\rvtheta^*\right\|_2^2 - \left\|\rvtheta^{t+1} -\rvtheta^* \right\|_2^2 \right) + 
     \eta_\theta \left\|g_\theta(\rvtheta^t, \rvx^{t+1})\right\|_2^2  + \\
    & \frac{1}{nh_x\eta_\theta}\left\|\rvx^t-\rvx^*\right\|_{\sqrt{\hat{V}^{t+1}}}^2 +
    \frac{1}{nh_x\eta_\theta}\left\|\rvx^{t+1} -\rvx^* \right\|_{\sqrt{\hat{V}^{t+1}}}^2 + \\
    & \frac{n-2}{n^2}h_x\eta_\theta\left\|g_x(\rvtheta^t, \rvx^t)\right\|_{(\hat{V}^{t+1})^{-1/2}}^2 + \\
    &\frac{L_x(h_x\eta_\theta)^2}{n^2}\left\|g_x(\rvtheta^t, \rvx^{t})\right\|_{(\hat{V}^{t+1})^{-1}}^2 \Bigg].
\end{aligned}
\end{equation}
Summing the inequality from $1$ to $T$, the regret $R(T) = R_\theta(T) + R_x(T)$ is then upper bounded by:
\begin{equation}
\begin{aligned}
    R_\theta(T) &\le \frac{1}{2\eta_\theta^*}\left\|\rvtheta_1-\rvtheta\right\|_2^2  + \frac{\eta_\theta}{2}\mathbb{E} \sum_{t=1}^T  \left\|\hat{g}_\theta^k(\rvtheta^t, \rvx^{t+1})\right\|_2^2 \\
    &\le \frac{D_{\theta, 2}^2}{2\eta_\theta} + \frac{T\eta_\theta G_{\theta,2}^2}{2},
\end{aligned}
\end{equation}
and 
\begin{equation}
\begin{aligned}
    R_x(T) 
    \le & \sum_{t=1}^T \Bigg[
    \frac{1}{n\eta_x}\left(\left\|\rvx^t-\rvx^*\right\|_{\sqrt{\hat{V}^{t+1}}}^2 - \left\|\rvx^{t+1} -\rvx^* \right\|_{\sqrt{\hat{V}^{t+1}}}^2\right)  + \\
    &\quad\quad\  \frac{n-2}{n^2}\eta_x\left\|g_x(\rvtheta^t, \rvx^t)\right\|_{(\hat{V}^{t+1})^{-1/2}}^2 + 
    \frac{L_x\eta_x^2}{n^2}\left\|g_x(\rvtheta^t, \rvx^{t})\right\|_{(\hat{V}^{t+1})^{-1}}^2 \Bigg].
\end{aligned}
\end{equation}
$R_\theta(T)$ is the same as SGD. Using the inequality of arithmetic and geometric means, the optimality is achieved by $\eta_\theta = \frac{D_{\theta, 2}}{G_{\theta,2}\sqrt{T}}$ and we then have: 
\begin{equation}
R_\theta(T) \le G_{\theta,2} D_{\theta,2} \sqrt{T}.
\end{equation}
For the first term in $R_x(T)$, we have:
\begin{equation}
\begin{aligned}
    &\sum_{t=1}^T\left(\left\|\rvx^t-\rvx^*\right\|_{\sqrt{\hat{V}^{t+1}}}^2 - \left\|\rvx^{t+1} -\rvx^* \right\|_{\sqrt{\hat{V}^{t+1}}}^2 \right)\\
    = & \sum_{i=1}^n \Big(\sum_{t=2}^T (\sqrt{\hat{v}^{t+1}_i} - \sqrt{\hat{v}^{t}_i}) \|\rvx^t_i - \rvx_i^*\|_2^2 + \sqrt{\hat{v}^2_i} \|\rvx^1_i - \rvx_i^*\|_2^2 \Big) \\
    = & \sum_{i=1}^n \sum_{j=1}^d\Big(\sum_{t=2}^T (\sqrt{\hat{v}^{t+1}_i} - \sqrt{\hat{v}^{t}_i}) (\rvx^t_{i,j} - \rvx_{i,j}^*)^2  + \sqrt{\hat{v}^2_i} (\rvx^1_{i,j} - \rvx_{i,j}^*)^2 \Big), \\
\end{aligned}
\end{equation}
where $\rvx_{i,j}$ means the $j$-th coordinate of $\rvx_i$. Since $\forall i,j, t, |\rvx^t_{i,j}-\rvx_{i,j}^*| < D_{x, \infty}$ is assumed, then
\begin{equation}
\begin{aligned}
    &\sum_{t=1}^T\left(\left\|\rvx^t-\rvx^*\right\|_{\sqrt{\hat{V}^{t+1}}}^2 - \left\|\rvx^{t+1} -\rvx^* \right\|_{\sqrt{\hat{V}^{t+1}}}^2 \right)\\
    \le & \sum_{i=1}^n \sum_{j=1}^d\left(\sqrt{\hat{v}^2_i} D_{x, \infty}^2 + \sum_{t=2}^T (\sqrt{\hat{v}^{t+1}_i} - \sqrt{\hat{v}^{t}_i}) D_{x, \infty}^2 \right) \\
    = & \sum_{i=1}^n \sum_{j=1}^d\left(\sqrt{\hat{v}^{T+1}_i} D_{x, \infty}^2 \right) \\
    \le & \sum_{i=1}^n \left(d D_{x, \infty}^2 \sqrt{\hat{v}^{T+1}_i}\right) \\
    \le & d D_{x,\infty}^2  \sum_{i=1}^n G_{x_i,2}.
\end{aligned}
\end{equation}
For the second term of $R_x(T)$, we have:
\begin{equation}
\begin{aligned}
    &\left\|g_x(\rvtheta^t, \rvx^t)\right\|_{(\hat{V}^{t+1})^{-1/2}}^2 \\
    = & \sum_{i=1}^n \sum_{j=1}^{d} \frac{(g_x^i(\rvtheta^t, \rvx^t)_j)^2}{\sqrt{v^{t+1}_i}} \\
    \le & \sum_{i=1}^n \sum_{j=1}^{d} \frac{(g_x^i(\rvtheta^t, \rvx^t)_j)^2}{\sqrt{1-\beta}\sqrt{\sum_{j=1}^{d} ((g_x^i(\rvtheta^t, \rvx^t)_j)^2}} \\
    = & \frac{1}{\sqrt{1-\beta}} \sum_{i=1}^n \left\|g_x^i(\rvtheta^t, \rvx^t)\right\|_2 \\
    \le & \frac{1}{\sqrt{1-\beta}} \sum_{i=1}^n G_{x_i,2},
\end{aligned}
\end{equation}
where $g_x^i(\rvtheta^t, \rvx^t)_j$ represents the $j$-th coordinate of $g_x^i(\rvtheta^t, \rvx^t)$. Summing over $t$, the second term of $R_x(T)$ is bounded by:
\begin{equation}
\begin{aligned}
    \sum_{t=1}^T\left\|g_x(\rvtheta^t, \rvx^t)\right\|_{(\hat{V}^{t+1})^{-1/2}}^2 
    \le \frac{T}{\sqrt{1-\beta}} \sum_{i=1}^n G_{x_i,2}.
\end{aligned}
\end{equation}
And the third term of $R_x(T)$ can be bounded by:
\begin{equation}
    \begin{aligned}
        &\left\|g_x(\rvtheta^t, \rvx^t)\right\|_{(\hat{V}^{t+1})^{-1}}^2 \\
        = & \sum_{i=1}^n \sum_{j=1}^{d} \frac{(g_x^i(\rvtheta^t, \rvx^t)_j)^2}{v^{t+1}_i} \\
        \le & \sum_{i=1}^n \sum_{j=1}^{d} \frac{(g_x^i(\rvtheta^t, \rvx^t)_j)^2}{(1-\beta)\sum_{j=1}^{d} ((g_x^i(\rvtheta^t, \rvx^t)_j)^2} \\
        = & \frac{1}{1-\beta}. \\
    \end{aligned}
\end{equation}
Therefore
\begin{equation}
\begin{aligned}
    \sum_{t=1}^T\left\|g_x(\rvtheta^t, \rvx^t)\right\|_{(\hat{V}^{t+1})^{-1}}^2 
    \le \frac{T}{1-\beta}.
\end{aligned}
\end{equation}
Combining these inequalities, $R_x(T)$ is bounded by:
\begin{equation}
\begin{aligned}
    R_x(T) \le & \frac{T\eta_x}{2n\sqrt{1-\beta}} \sum_{i=1}^n G_{x_i,2} + \frac{D_{x,\infty}^2 d}{2n\eta_x} \sum_{i=1}^n G_{x_i,2} + \frac{L_x \eta_x^2 T}{2n^2(1-\beta)}.
\end{aligned}
\end{equation}
Using the inequality of arithmetic and geometric means, the bound on the achieves the minimum when $\eta_x = \frac{\sqrt{d}D_{x, \infty}(1-\beta)^{1/4}}{\sqrt{T}}$ and
\begin{equation}
R_x(T) \le \frac{D_{x,\infty}\sum_{i=1}^n G_{x_i, 2} \sqrt{dT}}{n(1-\beta)^{1/4}} + \frac{dL_xD_{x,\infty}^2}{2n^2\sqrt{1-\beta}}.
\end{equation}
Combining $R_\theta(T)$ and $R_x(T)$, the regret of ASGDBCA is bounded by
\begin{equation}
\begin{aligned}
    R(T) \le & G_{\theta,2} D_{\theta,2} \sqrt{T} + \frac{D_{x,\infty}\sum_{i=1}^n G_{x_i, 2} \sqrt{dT}}{n(1-\beta)^{1/4}} + \\
    &\frac{dL_xD_{x,\infty}^2}{2n^2\sqrt{1-\beta}}.
\end{aligned}
\end{equation}
\end{proof}

\section{Additional Experiments}
\label{sec:addexp}
\subsection{Catastrophic Overfitting in ATTA}
\label{sec:addexp_atta}
As shown in \Cref{fig:loss_gap_atta}, when increasing the step size in ATTA, the loss gap between the ATTA and PGD10 becomes smaller. Furthermore, the robust accuracy also increases when the step size is not overly large in \Cref{fig:accuracy_stepsize_atta}. It shows that large step size also strengthen the attack in ATTA. However, large step also leads to catastrophic overfitting in ATTA. When the step size is overly large in \Cref{fig:accuracy_stepsize_atta}, the robust accuracy against PGD decreases to nearly 0\%, indicating catastrophic overfitting. 

As we show in the main text, instances with large gradient norm are more likely to cause catastrophic overfitting in ATTA. To verify it, we train a RN-18 with $\varepsilon=8/255$ on CIFAR10 with ATTA. We record the average gradient norm $GN(\rvx_i)$ and divide the subsets according to $\text{rank}(\rvx_i)$:
$$\gD_i^j = \{\rvx_k| \frac{10(i-1)}{n} \le \text{rank}(\rvx_k) < \frac{10j}{n}\}.$$ 
The training curves for different subsets are shown in \Cref{fig:co_atta}. It is the ATTA version of Figure 2 in the main text. When training with $\varepsilon=12/255\ \alpha=15/255$ or $\varepsilon=12/255\ \alpha=18/255$, the robust accuracy of the subsets of large gradient norm ($\gD_7^{10}, \gD_8^{10}, \gD_9^{10}$) suddenly decreases to nearly 0\%, indicating the phenomenon of  catastrophic overfitting. By contrast, catastrophic overfitting does not occur when training with the subsets of small gradient norm ($\gD_1^2, \gD_1^3, \gD_1^4$). The observation is the same as FGSM-RS in the main text.

\begin{figure}[t]
    \centering
    
	\begin{subfigure}[t]{0.35\linewidth}
	\centering
        \includegraphics[width=\linewidth]{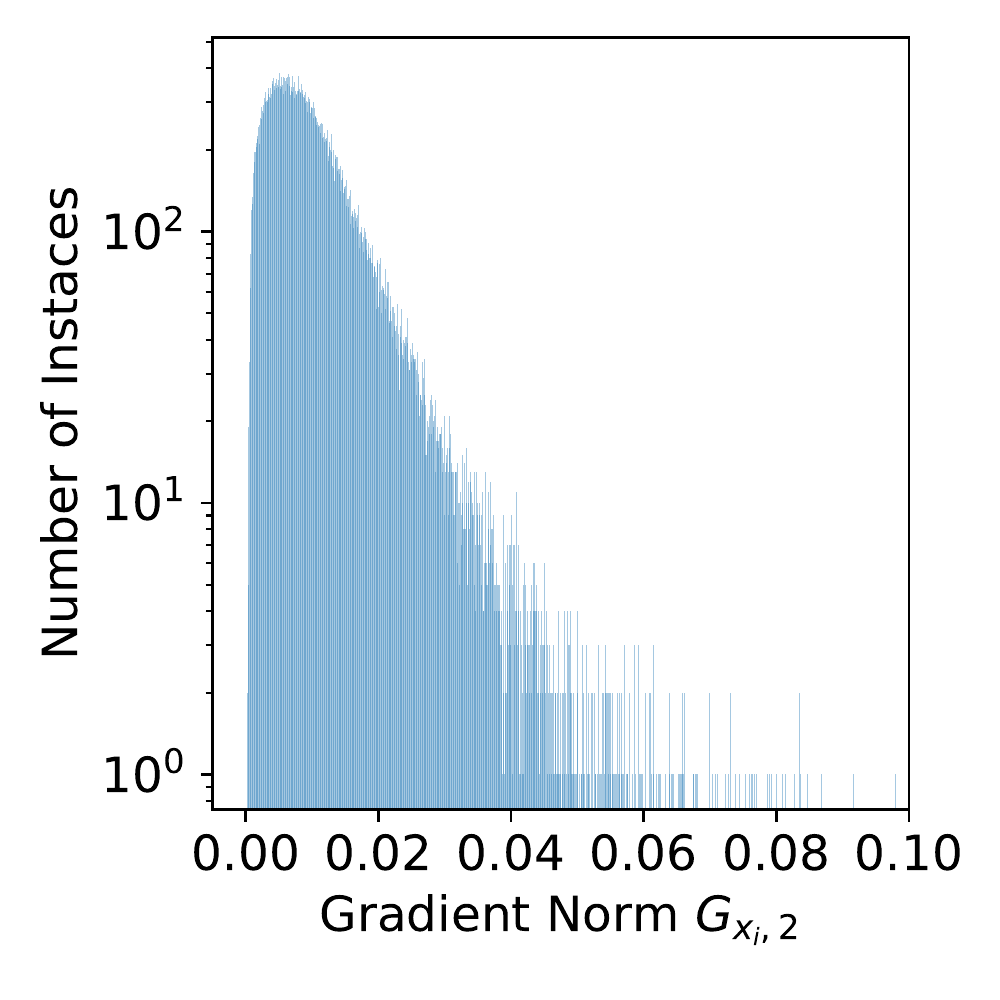}
        \captionsetup{font=normalsize}
        \vspace{-0.8cm}
        \caption{}
        \label{fig:gdnorm_hist}
    \end{subfigure}
	\begin{subfigure}[t]{0.45\linewidth}
	\centering
        \includegraphics[width=\linewidth]{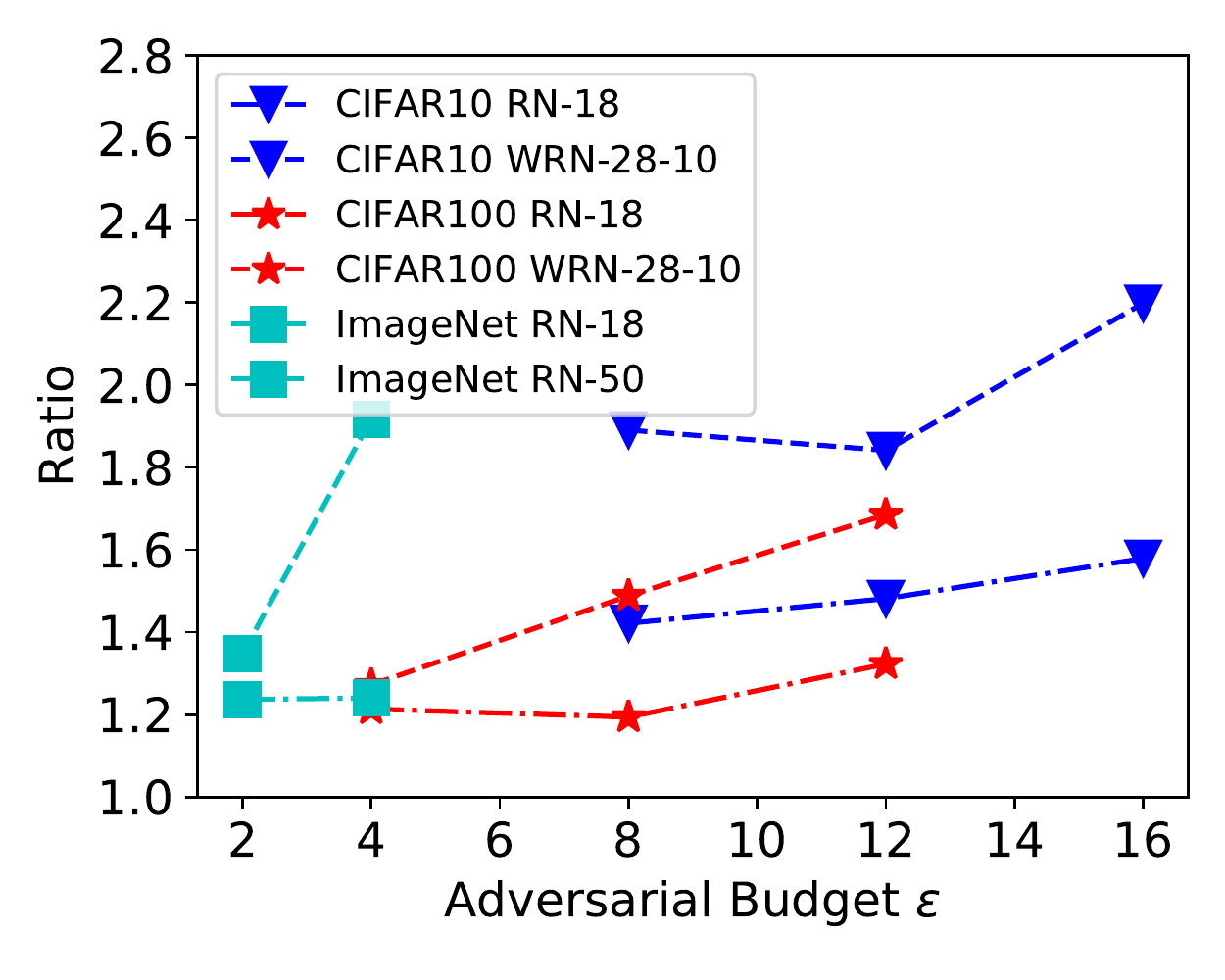}
        \captionsetup{font=normalsize}
        \vspace{-0.8cm}
        \caption{}
        \label{fig:gdnorm_ratio}
    \end{subfigure}
    \caption{(a) Histogram of $G_{x_i,2}$ of a ResNet-18 trained CIFAR10 with $\varepsilon=8/255$. (b) Ratio $\frac{\sum_{i=1}^n G_{x_i,2}^2}{n}\Big/(\frac{\sum_{i=1}^{n} G_{x_i,2}}{n})^2$ for different datasets and network architectures with different $\varepsilon$.}
    \label{fig:theory}
\end{figure}

\begin{figure}[t]
    \centering
	\begin{subfigure}[t]{0.4\linewidth}
	    \centering
        \includegraphics[width=\linewidth]{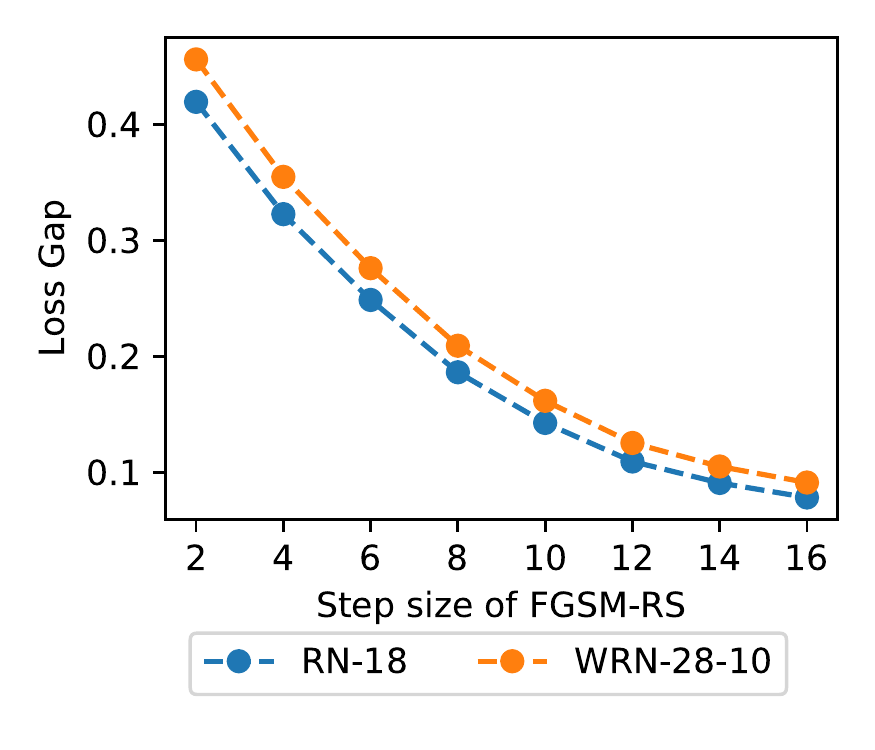}
        \captionsetup{font=normalsize}
        \vspace{-0.8cm}
        \caption{}
        \label{fig:loss_gap_atta}
    \end{subfigure}
	\begin{subfigure}[t]{0.4\linewidth}
	    \centering
        \includegraphics[width=\linewidth]{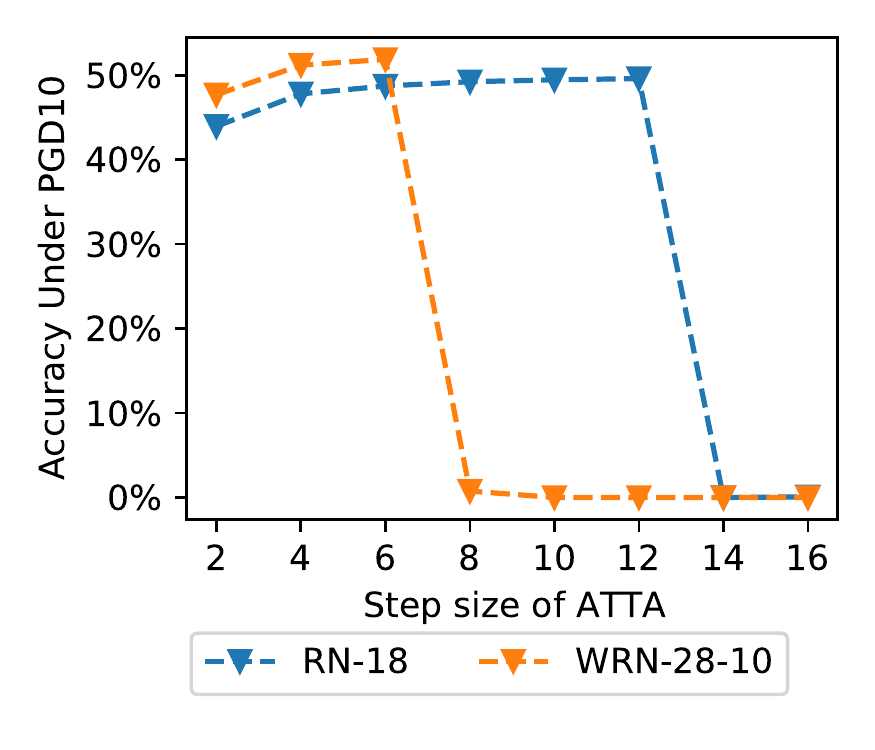}
        \captionsetup{font=normalsize}
        \vspace{-0.8cm}
        \caption{}
        \label{fig:accuracy_stepsize_atta}
    \end{subfigure}
    \caption{(a) The loss gap of training instances between PGD10 and ATTA attack $\ell(\rvx^{\text{PGD}}, y) - \ell(\rvx^{\text{FGSM-RS}}, y)$ with different step sizes for a ATTA trained robust model; (b) The test robust accuracy of the models trained by ATTA with different step sizes.}
\end{figure}

\begin{figure}[!t]
    \centering
    \includegraphics[width=0.4\linewidth]{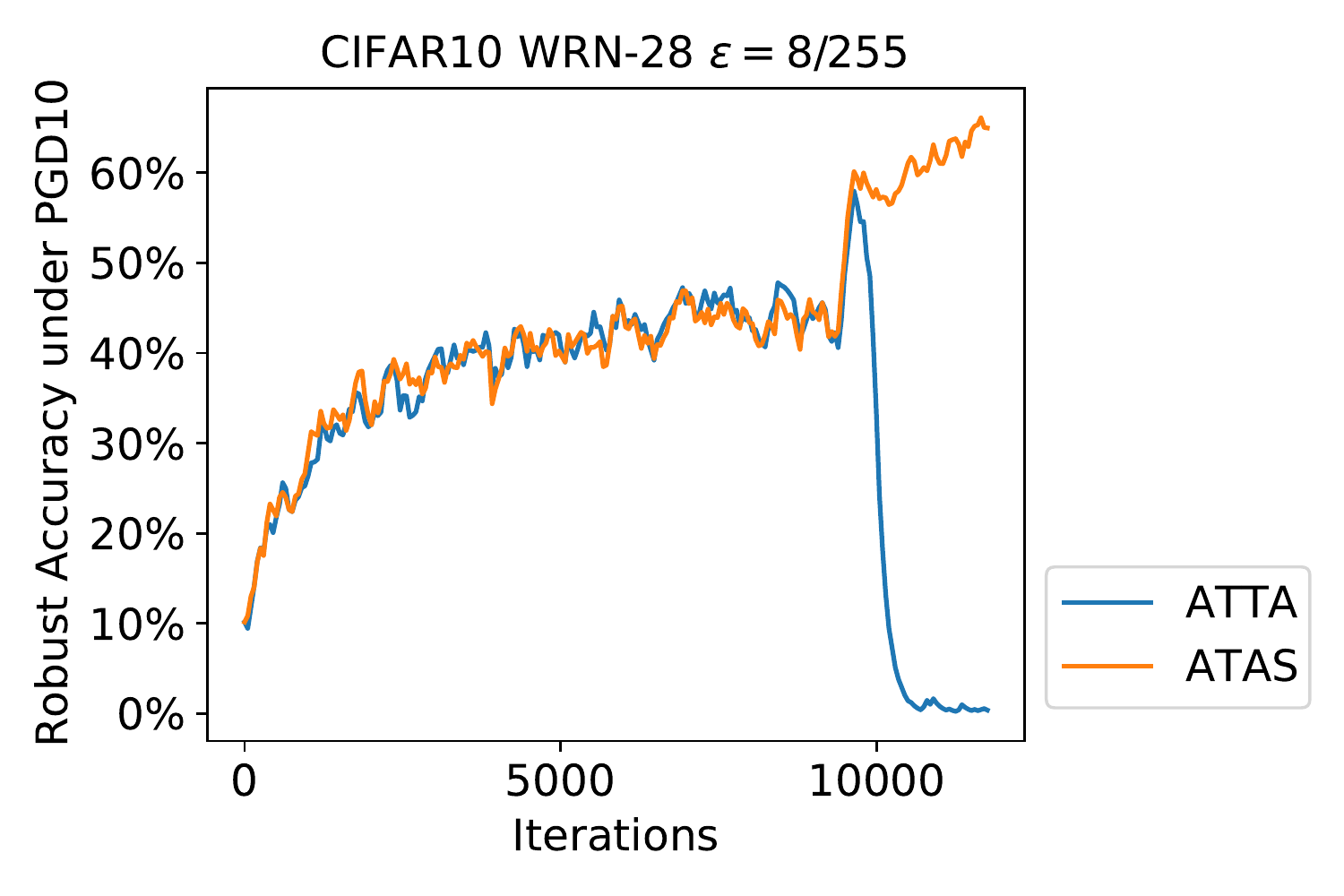}
    \vspace{-0.3cm}
    \caption{Training accuracy under PGD10 of ATTA ($\alpha=8/255$) and {\name}. Even if the average step size of {\name} ($\bar{\alpha}=9.3/255$) is larger than ATTA ($\alpha=8/255$), catastrophic overfitting does not occur in {\name}. }
    \label{fig:compare_atta_atas}

\end{figure}

\begin{figure}[t]
    \centering
    \includegraphics[width=\linewidth]{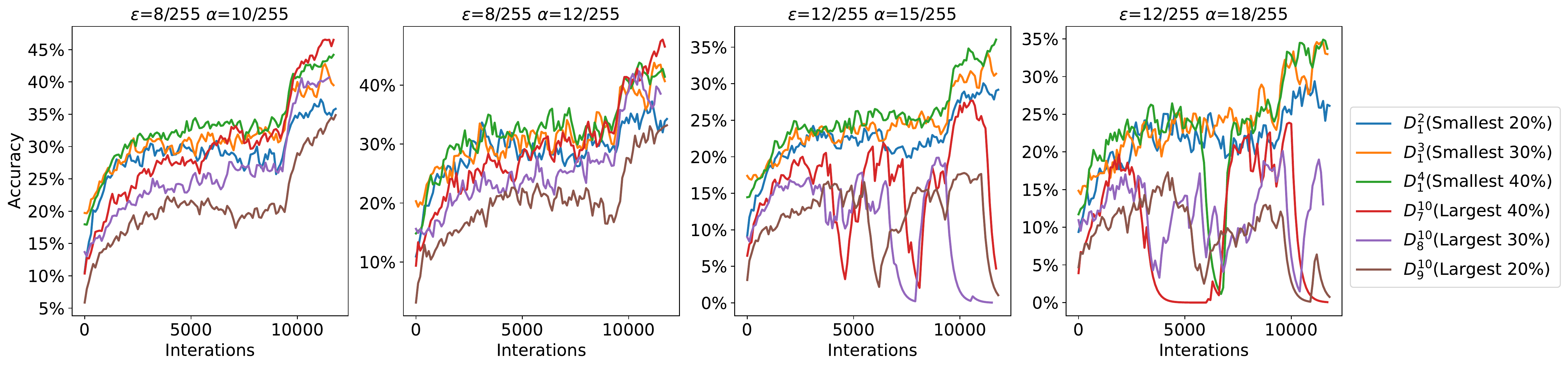}
    \vspace{-0.3cm}
    \caption{The robust training accuracy curve of FGSM-RS trained on different subsets of CIFAR10. The attack is PGD10 and the network is ResNet-18. The adversarial budgets and the step sizes are shown on top of each figure.}
    \label{fig:co_atta}
\end{figure}

\begin{table*}[t!]
    \caption{Robust accuracy and training time of different fast AT methods on CIFAR10 with $\varepsilon=8/255, 12/255, 16/255$.}
    \begin{subtable}{\linewidth}
        \centering
        \captionsetup{font=normalsize}
        \caption{CIFAR10 with adversarial budget $\varepsilon=8/255$.}
        \setlength\tabcolsep{3.4pt}
        \begin{tabular}{l c c c c c c c c c c}
            \hline
            \multirow{2}*{Methods} & \multicolumn{5}{c}{ResNet-18} & \multicolumn{5}{c}{WideResNet-28-10} \\
            \cmidrule(r){2-6}  \cmidrule(r){7-11}
    
            & Clean & PGD10 & PGD50 & AA & Time(h) & Clean & PGD10 & PGD50 & AA & Time(h) \\ \hline
            \textit{PGD10}& \textit{80.13} & \textit{50.59} & \textit{48.94}  & \textit{45.97} & \textit{1.23} & \textit{85.00} & \textit{55.51} & \textit{53.53}  & \textit{51.27} & \textit{8.49}\\         \hline
            FreeAT  & 78.37 & 40.90 &  39.02 &  36.00 & 0.33 & 84.54 & 46.09 &  43.80 &  41.19 & 2.31 \\
            YOPO  & 74.72 & 37.51 & 35.79 & 33.21 & 0.28 & 82.92 & 44.62 & 42.14 & 40.23 & 1.90 \\
            FGSM-RS& \textbf{83.99} & 48.99 & 46.36 & 42.95 & \textbf{0.22} & 80.21 & 0.01 & 0.00 & 0.00 & 1.67 \\
            FGSM-GA& 80.10 & 49.14 & 47.21 & 43.44 & 0.57 & 75.84 & 45.57 & 43.28 & 39.44 & 3.82 \\
            SSAT & 88.83 & 42.31 & 38.99 & 37.06 & 0.61 & 90.40 & 44.04 & 40.40 & 38.82 & 3.53 \\
            ATTA & 82.16 & 47.47 & 45.32 & 42.51 & 0.30 & 85.90 & 51.52 & 48.94 & 46.84 & 1.70\\
            {\name} & 81.22 & \textbf{50.03} & \textbf{48.18} & \textbf{45.38} & 0.30 & \textbf{85.96} & \textbf{53.43} & \textbf{51.03} & \textbf{48.72} & \textbf{1.63} \\ 
            \hline 
        \end{tabular}

        \vspace{0.3cm}
    \end{subtable}
    
    \begin{subtable}{\linewidth}
        \centering
        \captionsetup{font=normalsize}
        \caption{CIFAR10 with adversarial budget $\varepsilon=12/255$
        }
        \setlength\tabcolsep{3.4pt}
        \begin{tabular}{l c c c c c c c c c c}
            \hline
            \multirow{2}*{Methods} & \multicolumn{5}{c}{ResNet-18} & \multicolumn{5}{c}{WideResNet-28-10} \\
            \cmidrule(r){2-6}  \cmidrule(r){7-11}

            & Clean & PGD10 & PGD50 & AA & Time(h) & Clean & PGD10 & PGD50 & AA & Time(h) \\ \hline
            \textit{PGD10}& \textit{70.46} & \textit{39.39} & \textit{37.22}  & \textit{32.99} & \textit{1.23}  & \textit{76.49} & \textit{43.67} & \textit{40.93}  & \textit{36.95} & \textit{8.48} \\
            \hline 
            FreeAT  & 72.92 & 25.88 &  22.82 &  19.95 & 0.33 & 79.71 & 26.31 &  23.72 &  18.98 & 2.33 \\
            YOPO  & 64.21 & 23.82 & 22.27 & 17.16 & 0.29 & 75.29 & 32.27 & 28.41 & 25.42 & 1.92 \\
            FGSM-RS& \textbf{80.78} & 0.00 & 0.00 & 0.00 & \textbf{0.22} & 79.41 & 0.00 & 0.00 & 0.00 & 1.66 \\
            FGSM-GA& 68.62 & 36.76 & 33.96 & 28.57 & 0.57 & 72.87 & 37.98 & 35.18 & 29.01 & 3.82 \\
            SSAT & 89.08 & 6.50 & 1.16 & 0.03 & 0.60 & 91.45 & 0.07 & 0.00 & 0.00 & 3.50 \\
            ATTA & 74.46 & 35.85 & 31.69 & 27.85 & 0.28 & \textbf{80.05} &  38.29 & 34.01 & 29.85 & 1.63 \\
            {\name} & 72.58 & \textbf{38.10} & \textbf{35.58} & \textbf{30.56} & 0.29 & 78.16 & \textbf{41.88} & \textbf{38.94} & \textbf{33.58} & \textbf{1.62} \\ \hline
        \end{tabular}
        \vspace{0.3cm}

    \end{subtable}
    \begin{subtable}{\linewidth}
        \centering
        \captionsetup{font=normalsize}
        \caption{CIFAR10 with adversarial budget $\varepsilon=16/255$
        }
        \setlength\tabcolsep{3.4pt}
        \begin{tabular}{l c c c c c c c c c c}
            \hline
            \multirow{2}*{Methods} & \multicolumn{5}{c}{ResNet-18} & \multicolumn{5}{c}{WideResNet-28-10} \\
            \cmidrule(r){2-6}  \cmidrule(r){7-11}
            & Clean & PGD10 & PGD50 & AA & Time(h) & Clean & PGD10 & PGD50 & AA & Time(h) \\ \hline
            \textit{PGD10}&  \textit{61.08} & \textit{31.42} & \textit{29.37}  & \textit{23.34}  & \textit{1.32} & \textit{66.57} & \textit{35.91} & \textit{32.89}  & \textit{27.24} & \textit{8.65} \\
            \hline 
            FreeAT & 61.05 & 15.86 &  12.49 &  10.04 & 0.33 & 67.89 & 19.07 & 14.76 &  12.53 & 2.34 \\
            YOPO & 67.45 & 14.87 & 12.00 & 8.66 & 0.29 & 62.75 & 18.30 & 16.27 & 11.00 & 1.91 \\
            FGSM-RS& 67.75 &  0.00 & 0.00 & 0.00 & \textbf{0.22} & 60.21 & 0.00 & 0.00 & 0.00 & 1.67  \\
            FGSM-GA& 54.07 & 27.05 & 25.10 & 18.92 & 0.57 & 16.08 & 13.38 & 13.29 & 11.44 & 3.77  \\
            SSAT & 90.41 & 0.43 & 0.03 & 0.00 & 0.59 & 91.42 & 0.00 & 0.00 & 0.00 & 3.52 \\
            ATTA & 63.37 & 26.66 & 23.45 & 17.02 & 0.29 & \textbf{72.90} & 30.11 & 23.92 & 19.11 & \textbf{1.65} \\
            {\name} &  \textbf{64.11} &  \textbf{31.39} & \textbf{28.15} & \textbf{21.09} & 0.30  & 70.33 & \textbf{34.32} & \textbf{30.53} & \textbf{22.58} & 1.68 \\ \hline
        \end{tabular}
    \end{subtable}
    \label{tab:cifar10_app}
\end{table*}

Adaptive step sizes in {\name} allow larger step sizes without causing catastrophic overfitting. In \Cref{fig:compare_atta_atas}, we show the comparison between ATTA and {\name}. Even if {\name} has larger step size than ATTA, it does not suffer from catastrophic overfitting like ATTA.

\subsection{Robust Accuracy of Various Adversarial Budget}
\label{sec:addexp_acc}
The robust accuracy for CIFAR10 with $\varepsilon=8/255$,$12/255$,$16/255$, CIFAR100 with $\varepsilon=4/255$, $8/255$, $12/255$ and ImageNet with $\varepsilon=2/255$, $4/255$ are provided in \Cref{tab:cifar10_app}, \Cref{tab:cifar100_app} and \Cref{tab:imagenet_app} respectively. {\name} achieves the best robust accuracy in all these experiments with different datasets, network architectures and adversarial budgets.

\subsection{Steps size and Gradient Norm }
\label{sec:addexp_step}
\begin{figure}[t]
	\begin{subfigure}[t]{0.49\linewidth}
		\centering
            \includegraphics[width=\linewidth]{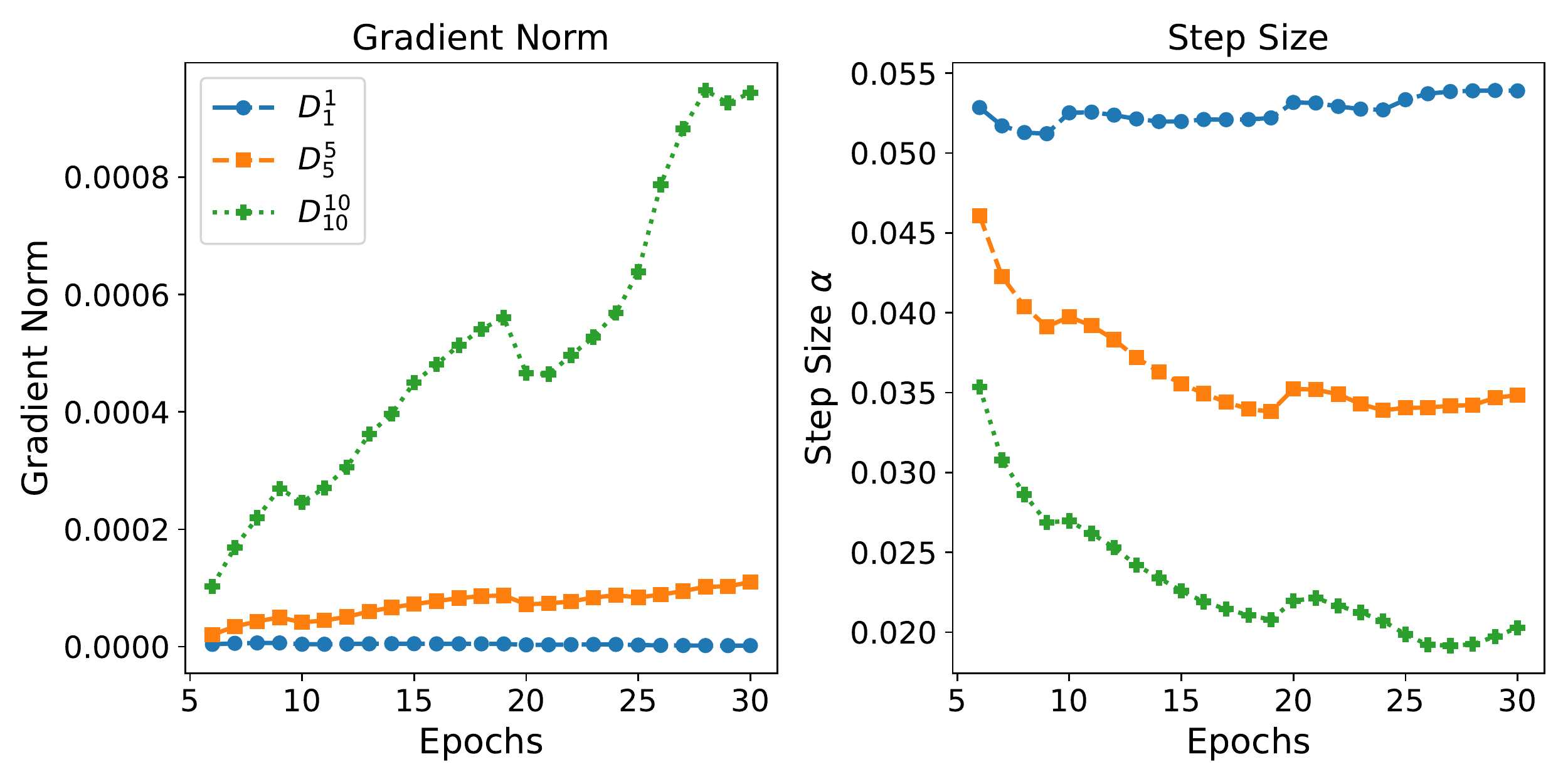}
            \caption{ResNet-18}
            \label{fig:step_gd_rn18}
    \end{subfigure}
	\begin{subfigure}[t]{0.49\linewidth}
		\centering
            \includegraphics[width=\linewidth]{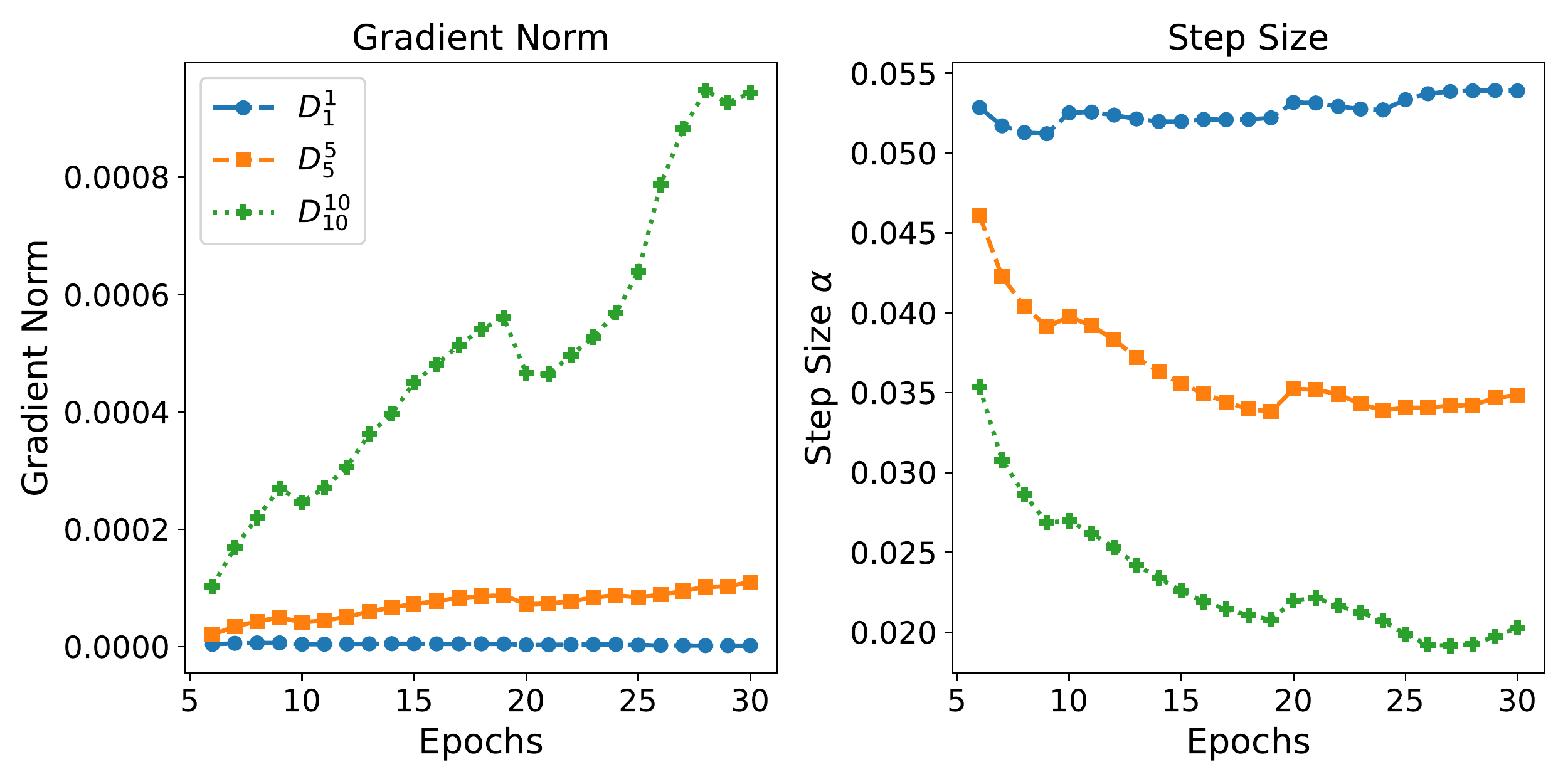}
            \caption{WideResNet-28-10}
            \label{fig:step_gd_wrn28}
    \end{subfigure}
    \caption{The input gradient norm and step size for CIFAR10 with $\varepsilon=8/255$ after warm up. }
    \label{fig:step_gd}
\end{figure}

\Cref{fig:step_gd} plots the changes of gradient norm and step size for {\name} after warm up. We divide CIFAR10 into 10 subsets according to their gradient norm and plot the gradient norm and step size for $\mathcal{D}_1^1$, $\mathcal{D}_5^5$ and $\mathcal{D}_{10}^{10}$, which has the smallest, medium and largest input gradient norm among 10 subsets. The figure shows that the input gradient norm and step size is relatively stable for each subset along the training process. It shows that the input gradient norm is more like a property of training instances themselves, which is consistent with our motivation. It is worth noting that the sudden changes of gradient norm is the result of initialization reset used in ATTA.

\subsection{Convergence Gap}
\label{sec:addexp_convergence}
In \Cref{tab:convergence_gap}, we show the relationship between the \textit{Ratio} 
$$
\text{Ratio} = \frac{1}{(1-\beta)^{\frac{1}{4}}}\sqrt{\frac{\sum_{i=1}^n G_{x_i,2}^2}{n}\Big/(\frac{\sum_{i=1}^{n} G_{x_i,2}}{n})^2}
$$
and the convergence gap $\ell_{\text{ATTA}} - \ell_{\text{ATAS}}$ and convergence ratio $\ell_{\text{ATTA}}/\ell_{\text{ATAS}}$ in the last epoch of training. Here, $\ell$ is the loss of each method. The ratio is obtained from \Cref{fig:gdnorm_ratio} for CIFAR10 with ResNet-18. It shows that larger \textit{Ratio} (more long-tailed distribution) leads to larger convergence gap between ATTA and {\name}. 
\begin{table}[t]
    \caption{Convergence gap and the ratio on CIFAR10 with ResNet-18.}
    \centering
    \setlength\tabcolsep{5.0pt}

    \begin{tabular}{c c c c}
        \hline
        Ratio & 1.4 ($\varepsilon$=8/255) & 1.5 ($\varepsilon$=12/255) & 1.6 ($\varepsilon$=16/255) \\ 
        Convergence Gap $\ell_{\text{ATTA}} - \ell_{\text{ATAS}}$ & 0.05 & 0.10 & 0.12  \\
        Convergence Ratio $\ell_{\text{ATTA}}/\ell_{\text{ATAS}}$ & 1.03 & 1.11 & 1.13  \\ \hline
    \end{tabular}
    \label{tab:convergence_gap}
\end{table}

\section{Details about the Experiments}
\label{sec:expsetting}

\subsection{Algorithms for {\name} in ImageNet}
\label{sec:atas_imagenet}
In the experiments of ATTA and {\name}, we utilize the local property of the adversarial examples \cite{huang2020corrattack, banditsTD} and only store the interpolated perturbation in the memory. We resize the perturbations from $224\times 224$ to $32\times 32$ for storage in the memory and up-sample it back when using it as the initialization for the next epoch. The detailed algorithm is shown in \Cref{alg:adaptatta_imagenet}. 

\begin{algorithm}[t] 
    \caption{{\name} for ImageNet}
    \label{alg:adaptatta_imagenet} 
    \begin{algorithmic}[1]
    \REQUIRE Training set $\gD$, The model $f_\rvtheta$ with loss function $\ell$, Adversarial budget $\varepsilon$, Hyperparameters $\gamma, \eta, c, N$
    \ENSURE Optimized model $f_{\rvtheta^*}$\\ 
    \STATE $v_i^0=0$ for $i=1, \cdots, n$
    \STATE $\delta_i^0 = $ Uniform($-\varepsilon, \varepsilon$) for $i=1, \cdots, n$
    \STATE Resize $\delta_i^0$ to $32\times 32$ for $i=1, \cdots, n$ and store them.
    \FOR{$ j = 1$ to $N$}
        \FOR {$\rvx_i, y_i \in \gD$}
            \STATE Resize $\delta_i^{j-1}$ to $224\times 224$
            \STATE $\rvx_i^{j-1}$ = $\rvx_i$ + $\delta_i^{j-1}$
            \STATE $v_i^{j} = \beta v_i^{j-1} + (1-\beta) \|\nabla_{\rvx_i^{j-1}} \ell(\rvx_i^{j-1}, y_i; \rvtheta)\|_2^2$
            \STATE $\alpha_i^{j} = \gamma/(c + \sqrt{v_i^j})$
            \STATE $\rvx_i^{j} = \Pi_{\gB_p(\rvx_i, \varepsilon)}[\rvx_i^{j-1} + \alpha_i^{j}\cdot \text{sgn} (\nabla_{\rvx_i^{j-1}}\ell(\rvx_i^j, y; \rvtheta))]$
            \STATE $\rvtheta = \rvtheta - \eta \nabla_{\rvtheta}\ell(\rvx_i^{j}, y; \rvtheta))$
            \STATE $\delta_i^j = \rvx_i^j - \rvx_i$
            \STATE Resize $\delta_i^j$ to $32\times 32$ and store it.
        \ENDFOR
    \ENDFOR
    \end{algorithmic}
\end{algorithm}

\begin{table*}[t!]
    \caption{Robust accuracy and training time of different fast AT methods on CIFAR100 with $\varepsilon=4/255, 8/255, 12/255$.}

    \begin{subtable}{\linewidth}
        \centering
        \caption{CIFAR100  with adversarial budget $\varepsilon=4/255$
        }
        \setlength\tabcolsep{3.4pt}
        \begin{tabular}{l c c c c c c c c c c}
            \hline
            \multirow{2}*{Methods} & \multicolumn{5}{c}{ResNet-18} & \multicolumn{5}{c}{WideResNet-28-10} \\
            \cmidrule(r){2-6}  \cmidrule(r){7-11}
            & Clean & PGD10 & PGD50 & AA & Time(h) & Clean & PGD10 & PGD50 & AA & Time(h) \\ \hline
            \textit{PGD10}&  \textit{63.22} & \textit{41.18} & \textit{40.57}  & \textit{37.75}  & \textit{1.22} & \textit{69.03} & \textit{45.27} & \textit{44.44}  & \textit{43.30} & \textit{8.61} \\
            \hline 
            FreeAT  & 54.38 & 32.21 &  31.68 &  28.26 & 0.32 & 63.91 & 39.39 & 38.64 & 35.52 & 2.29 \\
            YOPO & 59.04 & 34.55 & 34.02 & 31.45 & 0.28 & 64.60 & 38.92 & 38.27 & 35.39 & 1.89 \\
            FGSM-RS & \textbf{65.50} &  39.41 & 38.50 & 36.35 & \textbf{0.22} & 69.62 &  42.18 & 41.35 & 39.64 & \textbf{1.60} \\
            FGSM-GA & 57.33 & 35.49 & 35.01 & 32.03 & 0.57 & 68.45 & \textbf{51.92} & \textbf{44.09} & 41.03 & 3.80  \\
            SSAT & 70.81 & 33.17 & 31.09 & 29.81 & 0.60 & 74.43 & 36.51 & 34.55 & 33.34 & 3.52 \\
            ATTA & 64.28 & 39.55 & 39.20 & 36.03 & 0.29 & \textbf{69.51} &  \textbf{44.36} & 42.66 & 40.99 & 1.66 \\
            {\name} & 63.79 & \textbf{40.68} & \textbf{40.02} & \textbf{37.30} & 0.30 & 69.64 & 44.34 & 43.36 & \textbf{41.32} & 1.66  \\ \hline
        \end{tabular}
        \vspace{0.3cm}

    \end{subtable}

    \begin{subtable}{\linewidth}
        \centering
        \captionsetup{font=normalsize}
        \caption{CIFAR100  with adversarial budget $\varepsilon=8/255$
        }
        \setlength\tabcolsep{3.4pt}
        \begin{tabular}{l c c c c c c c c c c}
            \hline
            \multirow{2}*{Methods} & \multicolumn{5}{c}{ResNet-18} & \multicolumn{5}{c}{WideResNet-28-10} \\
            \cmidrule(r){2-6}  \cmidrule(r){7-11}
            & Clean & PGD10 & PGD50 & AA & Time(h) & Clean & PGD10 & PGD50 & AA & Time(h) \\ \hline
            \textit{PGD10}&  \textit{54.08} & \textit{28.03} & \textit{27.23}  & \textit{23.04}  & \textit{1.32} & \textit{60.04} & \textit{31.70} & \textit{30.67}  & \textit{27.11} & \textit{8.53} \\
            \hline 
            FreeAT  & 50.56 & 19.57 &  18.58 &  15.09 & 0.33 & 59.38 & 24.41 & 23.00 & 19.60 & 2.30 \\
            YOPO & 51.55 & 20.65 & 19.17 & 16.05 & 0.29 & 50.35 & 19.44 & 18.36 & 15.43 & 1.92 \\
            FGSM-RS & \textbf{59.35} & 26.40 & 24.29 & 19.73 & \textbf{0.21} & 51.83 & 0.00 & 0.00 & 0.00 & \textbf{1.60} \\
            FGSM-GA & 50.61 & 24.48 & 24.07 & 19.42 & 0.57 & 54.29 & 25.86 & 24.56 & 20.74 & 3.80   \\
            SSAT & 71.03 & 9.79 & 4.80 & 1.09 & 0.62 & 75.01 & 0.21 & 0.01 & 0.00 & 3.50 \\
            ATTA  & 57.21 & 25.76 & 24.90 & 21.03 & 0.28 & \textbf{63.04} & 28.93 & 27.18 & 24.42 & 1.63\\
            {\name} &  55.49 & \textbf{27.68} & \textbf{26.60} & \textbf{22.62} & 0.31 & 62.34 & \textbf{29.89} & \textbf{28.35} & \textbf{25.03} & 1.61  \\ \hline
        \end{tabular}

        \vspace{0.3cm}

    \end{subtable}
    
    \begin{subtable}{\linewidth}
        \centering
        \captionsetup{font=normalsize}
        \caption{CIFAR100  with adversarial budget $\varepsilon=12/255$
        }
        \setlength\tabcolsep{3.4pt}
        \begin{tabular}{l c c c c c c c c c c}
            \hline
            \multirow{2}*{Methods} & \multicolumn{5}{c}{ResNet-18} & \multicolumn{5}{c}{WideResNet-28-10} \\
            \cmidrule(r){2-6}  \cmidrule(r){7-11}
            & Clean & PGD10 & PGD50 & AA & Time(h) & Clean & PGD10 & PGD50 & AA & Time(h) \\ \hline
            \textit{PGD10}&  \textit{44.31} & \textit{20.41} & \textit{19.22}  & \textit{15.41}  & \textit{1.34} & \textit{50.30} & \textit{23.81} & \textit{22.55}  & \textit{18.13} & \textit{8.56}  \\
            \hline 
            FreeAT  & 41.05 & 11.85 &  10.67 &  8.33 & 0.32 & 46.54 & 14.95 & 13.07 & 10.64 & 2.30 \\
            YOPO & 44.69 & 10.52 & 9.20 & 7.41 & 0.29 & 54.13 & 13.19 & 11.76 & 9.68 & 1.92\\
            FGSM-RS & 32.78 & 0.00 & 0.00 & 0.00 & \textbf{0.22} & 38.74 & 0.00 & 0.00 & 0.00 & \textbf{1.60} \\
            FGSM-GA & 39.77 & 17.06 & 16.07 & 12.14 & 0.57 & 51.05 & 20.54 & 19.37 & 14.77 & 3.80  \\
            SSAT & 71.38 & 3.00 & 1.18 & 0.09 & 0.60 & 75.50 & 0.00 & 0.00 & 0.00 & 3.56 \\
            ATTA  & \textbf{50.55} &  18.58 & 16.59 & 12.97 & 0.28 & \textbf{56.46} & 20.82 & 18.17 & 15.24 & 1.63 \\
            {\name} &  47.14 & \textbf{19.73} & \textbf{18.39} & \textbf{14.41} & 0.31 & 53.70 & \textbf{22.53} & \textbf{20.95} & \textbf{16.27} & 1.63  \\ \hline
        \end{tabular}
    \end{subtable}
    \label{tab:cifar100_app}
\end{table*}

\subsection{Detailed Hyperparameters for the Experiments}

As we focus on fast AT, we reduce the training epochs like \cite{fgsmga, fastat}. For single-step methods FGSM-RS, FGSM-GA, ATTA and {\name}, the training lasts for 30 epochs on CIFAR10 and CIFAR100, and 90 epochs on ImageNet. For FreeAT and YOPO, we keep the number of the forward-backward passes the same as the single-step methods so that the total training time of these methods will be similar. We use two kinds of learning rate scheduler: piece-wise decay used in \cite{atta} and cyclic learning rate used in \cite{fastat}, and choose the best scheduler for each method. 

\noindent\textbf{FreeAT.} We use the default hyperparameters from \cite{freeat} except training epochs to make fair comparison between different methods. We select the best number of batch replaying from \cite{freeat}. For CIFAR10 and CIFAR100, we use Free-8 in their paper (Free-$m$ means the number of batch replaying is $m$) and train the network for 10 epochs. For ImageNet, we use Free-4 and train the network for 45 epochs. 

\noindent\textbf{YOPO.} We use YOPO-5-3 in~\cite{yopo} as it achieves the best performance. The training lasts for 12 epochs for CIFAR10 and CIFAR100. For ImageNet, the training lasts for 36 epochs to make the training time similar to other methods. Other hyperparameters are the same as the original paper~\cite{yopo}.

\noindent\textbf{FGSM-RS.} We directly download the code from the official repository \url{https://github.com/locuslab/fast_adversarial}. The training lasts for 30 epochs for CIFAR10 and CIFAR100, and 90 epochs for ImageNet. Following the hyperparameters in the paper, the step size $\alpha=1.25\varepsilon$. Other hyperparameters are the same as their paper.

\begin{table*}[t!]
    \caption{Robust accuracy and training time of different fast AT methods on ImageNet with $\varepsilon=2/255, 4/255$.}
    
    \begin{subtable}{\linewidth}
        \centering
        \captionsetup{font=normalsize}

        \caption{ImageNet with adversarial budget  $\varepsilon=2/255$
        }
        \setlength\tabcolsep{3.4pt}
        \begin{tabular}{l c c c c c c c c c c}
            \hline
            \multirow{2}*{Methods} & \multicolumn{5}{c}{ResNet-18} & \multicolumn{5}{c}{ResNet-50} \\
            \cmidrule(r){2-6}  \cmidrule(r){7-11}
            & Clean & PGD10 & PGD50 & AA & Time(h) & Clean & PGD10 & PGD50 & AA & Time(h) \\ \hline
            FreeAT  & 58.80 & 35.56 &  34.78 &  31.77 & \textbf{40.01} & 65.81 & 44.12 & 43.34 & 40.80 & \textbf{108.3} \\
            YOPO  & 47.69 & 28.50 & 28.10 & 25.22 & 48.22 & 55.68 & 33.46 & 32.19 & 29.56 & 111.8 \\
            FGSM-RS & 55.26 & 37.33 & 36.98 & 33.28 & 43.46 & 67.83 & 46.12 & 45.56 & 43.58 & 115.0 \\
            FGSM-GA & 37.01 & 24.15 & 24.05 & 19.98 & 182.7 & / & / & / & / & /   \\
            ATTA & 58.32 & 39.62 & 38.32 & 36.08 & 45.83 & 66.62 & 48.27 & 47.65 & 45.00 & 111.7 \\
            {\name} &  \textbf{61.20} & \textbf{40.84} & \textbf{39.86} & \textbf{37.25} & 45.70 & \textbf{69.10} & \textbf{49.05} & \textbf{48.05} & \textbf{46.01} & 120.4\\ \hline
        \end{tabular}
        \vspace{0.3cm}
    \end{subtable}
    
        \begin{subtable}{\linewidth}
        \centering
        \captionsetup{font=normalsize}
        \caption{ImageNet  with adversarial budget  $\varepsilon=4/255$
        }
        \setlength\tabcolsep{3.4pt}
        \begin{tabular}{l c c c c c c c c c c}
            \hline
            \multirow{2}*{Methods} & \multicolumn{5}{c}{ResNet-18} & \multicolumn{5}{c}{ResNet50} \\
            \cmidrule(r){2-6}  \cmidrule(r){7-11}
            & Clean & PGD10 & PGD50 & AA & Time(h) & Clean & PGD10 & PGD50 & AA & Time(h) \\ \hline
            FreeAT  & 56.99 & 20.75 &  18.86 &  15.90 & 40.18 & 64.25 & 27.95 & 25.46 & 22.40 & 109.6\\
            YOPO  & 33.72 & 13.36 & 13.01 & 10.30 & 48.33 & 37.62 & 14.77 & 13.37 & 11.83 & 111.9 \\
            FGSM-RS & 49.73 & 26.48 & 25.70 & 21.11 & \textbf{38.84} & 66.75 & 1.08 & 0.13 & 0.00 & \textbf{103.6} \\
            FGSM-GA & 29.34 & 15.58 & 15.42 & 10.94 & 180.2 & / & / & / & / & /  \\
            ATTA & 54.25 & 27.31 & 26.97 & 22.47 & 47.95 & 63.28 & 35.13 & 33.37 & 29.46 & 114.8 \\
            {\name} &  \textbf{55.69} & \textbf{29.23} & \textbf{28.13} & \textbf{24.13} & 44.15 & \textbf{65.26} & \textbf{35.74} & \textbf{33.58} & \textbf{30.07} & 110.7 \\ \hline
        \end{tabular}

    \end{subtable}
    \label{tab:imagenet_app}
\end{table*}

\noindent\textbf{FGSM-GA.} We directly download the code from the official repository \url{https://github.com/tml-epfl/understanding-fast-adv-training}. The training lasts for 30 epochs for CIFAR10 and CIFAR100, and 90 epochs for ImageNet. Other hyperparameters are the same. For the experiments not involved in their paper, we keep them same as the experiments of CIFAR10 except for the hyperparameter $\lambda$ balancing the gradient align regularizer, which also varies for different datasets and adversarial budgets in their code. $\lambda$ for CIFAR10 is provided in their code. For CIFAR100 and ImageNet, we run several experiments and provide the result with best $\lambda$. These $\lambda$ are provided in \Cref{tab:fgsm_ga_lambda}. 

\noindent\textbf{SSAT.} We directly download the code from the official repository \url{https://github.com/Harry24k/catastrophic-overfitting}. The training lasts for 30 epochs for CIFAR10 and CIFAR100. And we use the check points $c=3$, which achieves the best performance in their paper.

\begin{table}[h]
    \centering
    \caption{Hyperparameter $\lambda$ for FGSM-GA}
    \begin{subtable}{0.49\linewidth}
        \centering
        \setlength\tabcolsep{5.0pt}
        \captionsetup{font=normalsize}
        \caption{CIFAR100}
        \begin{tabular}{c c c c}
            \hline
            $\varepsilon$ & $4/255$ & $8/255$ & $12/255$ \\
            $\lambda$ &  0.2 &  0.5 & 1.0 \\ \hline
        \end{tabular}
    \end{subtable}
    \begin{subtable}{0.49\linewidth}
        \centering
        \setlength\tabcolsep{5.0pt}
        \captionsetup{font=normalsize}
        \caption{ImageNet}
        \begin{tabular}{c c c}
            \hline
            $\varepsilon$ & $2/255$ & $4/255$ \\
            $\lambda$ &  0.005 &  0.01\\ \hline
        \end{tabular}
    \end{subtable}
    \label{tab:fgsm_ga_lambda}
\end{table}

\noindent\textbf{ATTA.} We follow the hyperparameters setting for ATTA-1 in \cite{atta} and set the step size $\alpha=4/255$. We reduce the number of training epochs to 30 for CIFAR10 and CIFAR100. And the epochs of piece-wise learning rate are rescheduled accordingly. The learning rate $\eta$ starts at 0.1 and decays to 0.01 and 0.001 at the 24th and 28th epochs. The training of ImageNet lasts for 90 epochs and the learning rate also starts at 0.1 and decays to 0.01 and 0.001 at the 50th and 75th epochs. The weight decay is $5\times 10^{-4}$ for CIFAR10 and CIFAR100. For ImageNet, it is $1\times10^{-4}$. The batch size is 128 for all the experiments. Other hyperparameters are the same as their paper.

\noindent\textbf{\name.} The hyperparameters $\gamma$ and $c$ are used to control the minimum and maximum step size for the training instances. \change{When the moving average of gradient norm $v_i^j \rightarrow 0 $, the step size $\alpha_i^j = \gamma/c$. We choose $\gamma/c = 16/255$, which is close to the adversarial budget. And $c$ should be close to the magnitude of $v_i^j$. As the gradient norm increases with the dimension of the inputs, $c$ should be larger for ImageNet. Therefore, we set $c=0.01$ and for CIFAR10 and CIFAR100 and we let $c=0.1$ for ImageNet. Momentum of gradient norm $\beta$ is set to 0.5 for all the experiments. {\name} is not sensitive to the choice of hyperparameters. Other hyperparameters are the same as ATTA. } 

\subsection{Environments of the Experiments}

All the training time is evaluated on a machine with \textit{Intel Xeon 8255C} and \textit{NVIDIA Tesla V100}. For CIFAR10 and CIFAR100, we use a single GPU. For ImageNet, we use two GPUs. We run all the experiments with \textit{Pytorch 1.4}.

\end{document}